\documentclass[fleqn,10pt]{wlscirep}
\newtheorem{mydef}{Definition}
\usepackage{adjustbox}
\usepackage{color,soul}
\usepackage{graphicx}
\usepackage{caption}
\usepackage{subcaption}
\usepackage{bm}
\usepackage{csquotes}
\usepackage{multirow}
\usepackage{todonotes}
\providecommand{\e}[1]{\ensuremath{\times 10^{#1}}}

\title{Formal Definitions of Unbounded Evolution and Innovation Reveal Universal Mechanisms for Open-Ended Evolution in Dynamical Systems}

\author[1,2,3]{Alyssa Adams}
\author[3,4,5]{Hector Zenil}
\author[1]{Paul C.W. Davies}
\author[1,6,7,8,*]{Sara Imari Walker}

\affil[1]{Beyond Center for Fundamental Concepts in Science, Arizona State University, Tempe AZ USA}
\affil[2]{Department of Physics, Arizona State University, Tempe AZ USA}
\affil[3]{Algorithmic Nature Group, LABORES, Paris, France}
\affil[4]{Department of Computer Science, University of Oxford UK}
\affil[5] {Unit of Computational Medicine, SciLifeLab, Department of Medicine Solna, Karolinska Institute, Stockholm Sweden}
\affil[6]{School of Earth and Space Exploration, Arizona State University, Tempe AZ USA}
\affil[7]{ASU-SFI Center for Biosocial Complex Systems, Arizona State University, Tempe AZ USA}
\affil[8]{Blue Marble Space Institute of Science, Seattle WA USA}

\affil[*]{To whom correspondence should be addressed: sara.i.walker@asu.edu}


\begin{abstract}
Open-ended evolution (OEE) is relevant to a variety of biological, artificial and technological systems, but has been challenging to reproduce {\it in silico}. Most theoretical efforts focus on key aspects of open-ended evolution as it appears in biology. We recast the problem as a more general one in dynamical systems theory, providing simple criteria for open-ended evolution based on two hallmark features: unbounded evolution and innovation. We define unbounded evolution as patterns that are non-repeating within the expected Poincar\'e recurrence time of an equivalent isolated system, and innovation as trajectories {\it not} observed in isolated systems. As a case study, we implement novel variants of cellular automata (CA) in which the update rules are allowed to vary with time in three alternative ways. Each is capable of generating conditions for open-ended evolution, but vary in their ability to do so. We find that \textit{state-dependent} dynamics, widely regarded as a hallmark of life, statistically out-performs other candidate mechanisms, and is the only mechanism to produce open-ended evolution in a scalable manner, essential to the notion of ongoing evolution. This analysis suggests a new framework for unifying mechanisms for generating OEE with features distinctive to life and its artifacts, with broad applicability to biological and artificial systems.
\end{abstract}
\begin{document}

\flushbottom
\maketitle
\thispagestyle{empty}

\section*{Introduction}

Many real-world biological and technological systems display rich dynamics, often leading to increasing complexity over time that is limited only by resource availability. A prominent example is the evolution of biological complexity: the history of life on Earth has displayed a trend of continual evolutionary adaptation and innovation, giving rise to an apparent open-ended increase in the complexity of the biosphere over its $> 3.5$ billion year history \cite{bedau1998classification}. Other complex systems, from the growth of cities \cite{bettencourt2007growth}, to the evolution of language \cite{seyfarth2005primate}, culture \cite{buchanan2011measuring, skusa2003towards} and the Internet \cite{okaopen} appear to exhibit similar trends of innovation and open-ended dynamics. Producing computational models that generate sustained patterns of innovation over time is therefore an important goal in modeling complex systems as a necessary step on the path to elucidating the fundamental mechanisms driving open-ended dynamics in both natural and artificial systems. If successful, such models hold promise for new insights in diverse fields ranging from biological evolution to artificial life and artificial intelligence.

Despite the significance of realizing open-ended evolution in theoretical models, progress in this direction has been hindered by lack of a universally accepted definition for {\it open-ended evolution} (OEE). Although relevant to many fields, OEE is most often discussed in the context of artificial life, where the problem is so fundamental that it has been dubbed a ``millennium prize problem'' \cite{bedau2000open}. Many working definitions exist, which can be classified into four hallmark categories as outlined by Banzhaf {\it et al.} \cite{banzhaf2016defining}: (1) on-going innovation and generation of novelty \cite{taylor1999artificial, ruiz2008enabling}; (2) unbounded evolution \cite{bedau1998classification, bedau1991can, packard1}; (3) on-going production of complexity \cite{fernando2011evolvability, ruiz2012autonomy, guttenberg2008cascade}; (4) a defining feature of life \cite{ruiz2004universal}. Each of these faces its own challenges, as each is cast in terms of equally ambiguous concepts. For example, the concepts of  ``innovation'' or ``novelty'', ``complexity'' and ``life'' are all notoriously difficult to formalize in their own right. It is also not apparent whether ``unbounded evolution'' is physically possible since real systems are limited in their dynamics by finite resources, finite time, and finite space. A further challenge is identifying whether the diverse concepts of OEE are driving at qualitatively different phenomena, or whether they might be unified within a common conceptual framework. For example, it has been suggested that increasing complexity might not itself be a hallmark of OEE, but instead a consequence of it \cite{taylor1999artificial, ruiz2004universal}. Likewise, processes may appear unbounded, even within a finite space, if they can continually produce novelty within observable dynamical timescales \cite{taylor2016open}. 

Given these limitations, it was unclear if OEE is a property unique to life, is inclusive of its artifacts (such as technology), or if it is an even broader phenomenon that could be a universal property of certain classes of dynamical systems. Many approaches aimed at addressing the hallmarks of OEE have been inspired by biology \cite{taylor2016open}, primarily because biological evolution is the best known example of a real-world system with the potential to be truly open-ended \cite{bedau1998classification}. However, as stated, other examples of potentially open-ended complex systems do exist, such as trends associated with cultural \cite{buchanan2011measuring, skusa2003towards} and technological \cite{bettencourt2007growth, okaopen} growth, and other creative processes.  Therefore, herein we set out to develop a more general framework to seek links between the four aforementioned hallmarks of OEE within dynamical systems, while remaining agnostic about their precise implementation in biology. Our motivation is to discover {\it universal mechanisms} that underlie OEE as it might occur both within and outside of biological evolution. 

In dynamical systems theory there exists a natural bound on the complexity that can be generated by a finite deterministic process, which is given by the \textit{Poincar\'e recurrence time}. Roughly, the Poincar\'e time is the maximal time after which any finite system returns to its initial state and its dynamical trajectory repeats. Clearly, new dynamical patterns cannot occur past the Poincar\'e time if the system is isolated from external perturbations. To cast the concept of unbounded evolution firmly within dynamical systems theory, we introduce a formal minimal criteria for {\it unbounded evolution} (where we stress that here we mean the broader concept of dynamical evolution, not just evolution in the biological sense) in finite dynamical systems: minimally, an unbounded system is one that does {\it not repeat} within the expected Poincar\'e time. A key feature is that this definition automatically excludes finite deterministic systems unless they are open to external perturbations in some way. That is, we contend that unbounded evolution (and in turn OEE which depends on it) is only possible for subsystem interacting with an external environment. To make better contact with real-world systems, where the Poincar\'e time often cannot even in principle be observed, we introduce a second criteria of {\it innovation}. Systems satisfying the minimal definition of unbounded evolution must also satisfy a formal notion of {\it innovation}, where we define innovation as dynamical trajectories {\it not} observed in isolated, unperturbed systems. We identify innovation by comparison to counterfactual histories (those of isolated systems). Like unbounded evolution,  innovation is extrinsically defined and requires interaction between at least two subsystems. A given subsystem can exhibit OEE {\it if only if} it is both {\it unbounded} and {\it innovative}. As we will show, utilizing this criteria for OEE allows us to evaluate candidate mechanisms for generating OEE in simple toy model dynamical systems, ones that could carry over to more realistic complex dynamical systems. 

The utility of these definitions is that they provide a simple way to quantify intuition regarding hallmarks (1) and (2) of OEE for systems of {\it finite} size, which is applicable to {\it any} comparable dynamical system. They therefore provide a means to quantitatively evaluate, and therefore directly compare, different potential mechanisms for generating OEE. We apply these definitions to test three new variants of cellular automata (CA) for their capacity to generate OEE. A key feature of the new variants introduced is their implementation of {\it time-dependent update rules}, which represents a radical departure from more traditional approaches to dynamical systems where the dynamical laws remain fixed. Each variant introduced differs in its relative openness to an external environment. Of the variants tested, our results indicate that systems that implement time-dependent rules that are a function of their state are statistically better at satisfying the two criteria for OEE than dynamical systems with externally driven time-dependence for their rules (that is where the rule evolution is not dependent on the state of the subsystem of interest). We show that the state-dependent systems provide a mechanism for generating OEE that includes the capacity for on-going production of novelty by coupling to larger environments. This mechanism is also scales with system size, meaning the amount of open-endedness that is generated does not drop off as the system size increases.  We then explore the complexity of state-dependent systems in more depth, calculating general complexity measures including compressibility (based on LZW in~\cite{zenil2010compression}) and Lyapunov exponents. Given that state-dependent dynamics are often cited as a hallmark feature of life due to the role of self-reference in biological processes \cite{alglife, davies2016hidden, goldenfeld2011life, douglas1979godel}, our results provide a new connection between hallmarks (1), (2) and (4) of OEE. Our results therefore connect several hallmarks of OEE in a new framework that allows identification of mechanisms that might operate in a diverse range of dynamical systems. The framework holds promise for providing insights into universal mechanisms for generating OEE in dynamical systems, which is applicable to both biological and artificial systems.

\section*{Theory} \label{sec:theory}

Traditionally dynamical systems, like their physical counterparts, are modeled with fixed dynamical laws -- a legacy from the time of Newton. However, this framework may not be the appropriate one for modeling biological complexity, where the dynamical laws appear to be self-referential and evolve in time as a function of the states \cite{alglife, davies2016hidden, goldenfeld2011life, douglas1979godel}. An explicit example is the feedback between genotype and phenotype within a cell: genes are ``read-out'' to produce changes to the state of expressed proteins and RNAs, and these in turn can feedback to turn individual genes on and off \cite{noble2012theory}.  Given this connection to biology, we are motivated in this work to focus explicitly on {\it time-dependent rules}, where time-dependence is introduced by driving the rule evolution through coupling to an external environment. Since open-ended evolution has been challenging to characterize in traditional models with fixed dynamical rules, implementing time-dependent rules could open new pathways to generating complexity. In this study we therefore define {\it open} systems as those where the rule dynamically evolves as a function of time, and we assume this is driven by interaction with an environment. As we show, time-dependent rules allow novel trajectories to be realized that have not been previously characterized in cellular automata models. To quantify this novelty, we introduce a rigorous notion of OEE that relies on formalized definitions of unbounded evolution and innovation. The definitions presented rely on utilizing isolated systems evolved according to a fixed rule as a set of counterfactual systems to compare to the novel dynamics driven by time-dependent rules. 

\subsection*{Formalizing Open-Ended Evolution as Unbounded Evolution and Innovation} \label{sec:Def}

\begin{figure}
\centering
        \centering
        \includegraphics[width=0.8\linewidth]{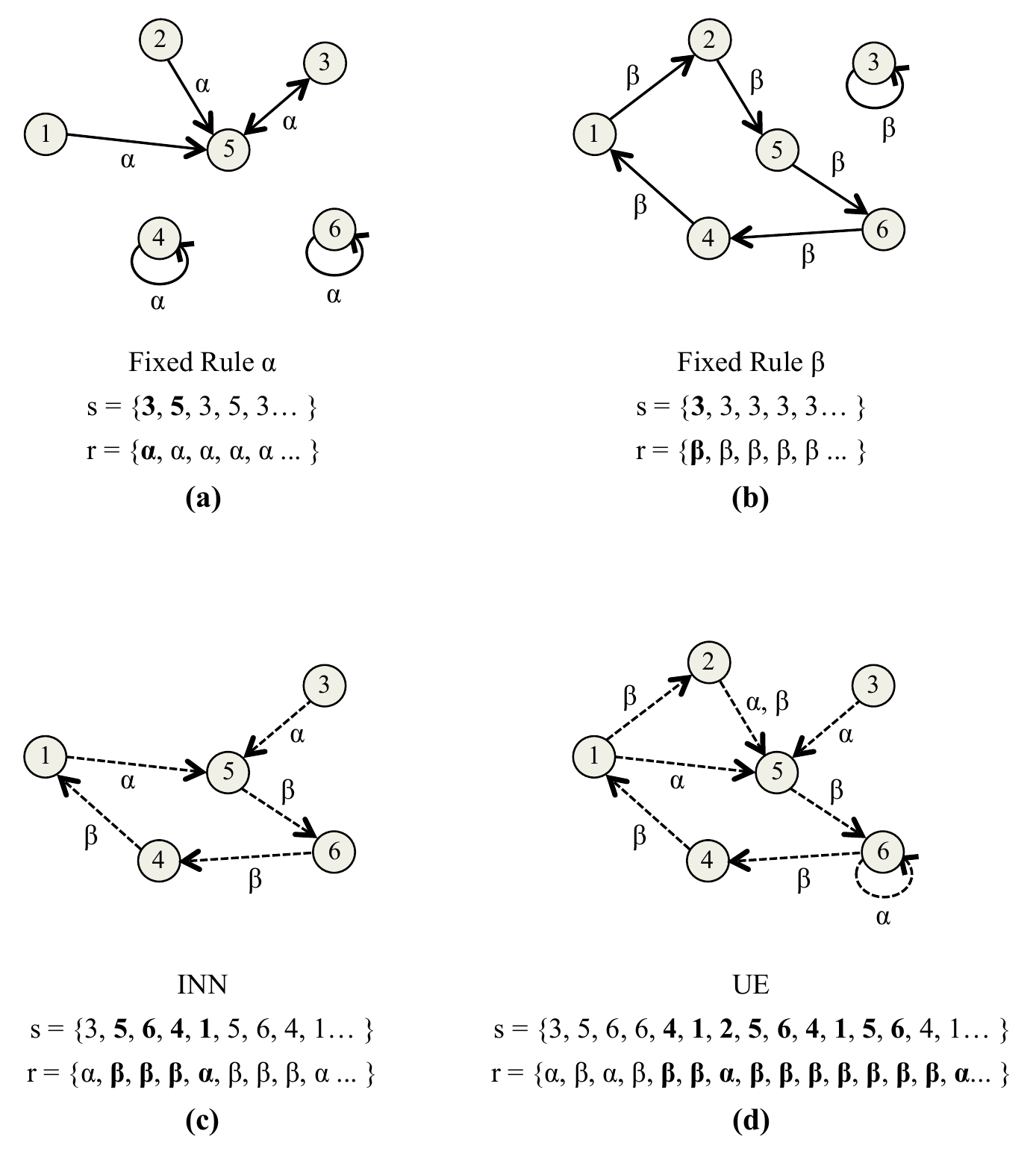}
\caption{State diagrams: A hypothetical example demonstrating the concepts of INN and UE. The possible set of states are $S = \{1, 2, 3, 4, 5, 6\}$ and rules $R = \{ \alpha, \beta \}$. For each panel, the example state trajectory $s$ is initialized with starting state $s_o = 3$. For panels \textbf{c} and \textbf{d} the rule trajectory $r$ is also shown. Highlighted in bold is the first iteration of the attractor for states (all panels) or rules  (panel \textbf{c} and \textbf{d} only). For a discrete deterministic system of six states, the Poincar\'e recurrence time is $t_P = 6$. Panel (\textbf{a}) shows the state transition diagram for hypothetical rule $\alpha$ where a trajectory initialized at $s (t_0) = 3$ visits two states. Panel (\textbf{b}) shows the state transition diagram for hypothetical rule $\beta$ where a trajectory initialized at $s (t_0) = 3$ visits only one state. Since the trajectories in (\textbf{a}) and (\textbf{b}) evolve according to a fixed rule (are isolated) they do not display INN or UE and in general the recurrence time $t_r \ll t_P$. Panel (\textbf{c}) demonstrates {\bf INN}, where the trajectory shown cannot be fully described by rule $\alpha$ or rule $\beta$ alone. The state trajectory $s$ and rule trajectory $r$ both have a recurrence time of $t_r = 5$, which is less than $t_P$ so this example does not exhibit UE. Panel (\textbf{d}) exhibits {\bf UE} (and is also an example of INN). The trajectory shown cannot be described by rule $\alpha$ or rule $\beta$ alone. The recurrence time for the state trajectory is $t_r = 13$, which is greater than $t_P$. The rule trajectory also satisfies the conditions for UE, with a recurrence time in this example that is longer than that of the state trajectory due to the fact that the state transition $2 \rightarrow 5$ could be driven by rule $\alpha$ or $\beta$ depending on the coupling to an external system.} \label{fig:INNUE}
\end{figure}

A hallmark feature of open-ended evolutionary systems is that they appear unbounded in their dynamical evolution \cite{bedau1998classification, bedau1991can, packard1}. For finite systems, such as those we encounter in the real world, the concept of ``unbounded'' is not well-defined. In part this is because all finite systems will eventually repeat, as captured by the well-known Poincar\'e recurrence theorem. As stated in the theorem, finite systems are bounded by their {\it Poincar\'e recurrence time}, which is the maximal time after which a system will start repeating its prior evolution. The Poincar\'e recurrence time $t_P$ of a finite, closed deterministic dynamical system therefore provides a natural bound on when one should expect such a system to stop producing novelty. In other words, $t_p$ is an absolute upper-bound on when such a system will terminate any appearance of open-endedness. 

Potentially the Poincar\'e recurrence theorem can {\it locally} be violated by a subsystem with open boundary conditions or if the subsystem is stochastic (although in the latter case the system might still be expected to approximately repeat). We therefore consider a definition of unbounded evolution applicable to {\it any} instance of a dynamical system that can be decomposed into two interacting subsystems. Nominally, we refer to these two interacting subsystems as the ``organism'' ($o$) and ``environment'' ($e$). We note that our framework is sufficiently general to apply to systems outside of biology: the concept of ``organism'' is meant only to stress that we expect this subsystem to potentially exhibit the rich dynamics intuitively anticipated of OEE when coupled to an environment (the environment, by contrast, is not expected to produce OEE behavior). The purpose of the second ``environment'' subsystem is to explicitly introduce external perturbations to the organism, where $e$ is also part of the larger system under investigation and modulates the rule of $o$ in a time-dependent manner. We therefore minimally define {\it unbounded evolution} (UE) as occurring when a sub-partition of a dynamical system does not repeat within its expected Poincar\'e recurrence time, giving the appearance of unbounded dynamics for given finite resources:

\begin{mydef} \label{Def:UE}
{\bf Unbounded evolution (UE)}: A system $U$ that can be decomposed into two interacting subsystems $o$ and $e$, exhibits unbounded evolution if there exists a recurrence time such that the state-trajectory or the rule-trajectory of $o$ is non-repeating for $t_r > t_P$ or $t_r' > t_P$ respectively, where $t_r$ is the recurrence time of the states, $t_r'$ the recurrence time of the rules, and $t_P$ is the Poincar\'e recurrence time for an equivalent isolated (non-perturbed) system $o$.
\end{mydef}

Since we consider $o$ where the states {\it and} rules evolve in time, unbounded evolution can apply to the state or rule trajectory recurrence time and still satisfy Definition \ref{Def:UE}. That is, a dynamical system exhibits UE {\it if and only if} it can be partitioned such that the sequence of one of its subsystems' states {\it or}  dynamical rules are {\it non-repeating} within the expected Poincar\'e recurrence time $t_P$ of an equivalent isolated system. In other words, unbounded evolution is only possible in a system that is partitioned into at least two interacting subsystems. This way, one of the subsystems acts as an external driver for the rule evolution of the other subsystem, which can then be pushed past its expected maximal recurrence time, $t_P$. We calculate the expected $t_P$ as that of an equivalent isolated system. By {\it equivalent isolated system}, we mean the set of {\it all} possible trajectories evolved from any initial state drawn from the same set possible of states as for $o$, but generated with a fixed rule, which can be any possible fixed rule.  We will describe explicit examples using the Elementary Cellular Automata (ECA) rule space in Section {\it Model Implementation}, where the relevant set of states are those constructed from the binary alphabet $\{0,1\}$ and the set of rules for comparison are the ECA rules. ECA are defined as 1-dimensional CA with nearest-neighbor update rules: for an ECA of width $w$ (number of cells across), equivalent isolated systems as defined here include all trajectories evolved with {\it any} fixed ECA rule from {\it any} initial state of width $w$, where $t_P$ is then $t_P = 2^w$ and $w=w_o$, where $w_o$ is the width of $o$.

Implementing the above definition of UE necessarily depends on counterfactual histories of {\it isolated} systems ({\it e.g.} of ECA in our examples). These counterfactual systems cannot, by definition, generate conditions for UE. This suggests as a corollary a natural definition for innovation in terms of comparison to the same set of counterfactual histories:

\begin{mydef} \label{Def:INN}
{\bf Innovation (INN)}: A system $U$ that can be decomposed into two interacting subsystems $o$ and $e$ exhibits innovation if there exists a recurrence time $t_r$ such that the state-trajectory is not contained in the set of all possible state trajectories for an equivalent isolated (non-perturbed) system. 
\end{mydef}

That is, a subsystem $o$ exhibits INN by Definition \ref{Def:INN} if its dynamics are {\it not} contained within the set of all possible trajectories of equivalent isolated systems. We note these definitions do not necessitate that the complexity of individual states increase with time, thus one might observe INN without a corresponding rise in complexity with time. Fig. \ref{fig:INNUE} shows a conceptual illustration of both UE and INN, as presented in Definitions \ref{Def:UE} and \ref{Def:INN}. 

A motivation for including both Definitions \ref{Def:UE} and \ref{Def:INN} is that they encompass intuitive notions of ``on-going production of novelty'' (INN) and ``unbounded evolution'' (UE), both of which are considered important hallmarks of OEE  \cite{banzhaf2016defining}.  UE can imply INN, but INN does not  likewise imply UE. It might therefore appear that UE is sufficient to characterize OEE without needing to appeal to separately defining INN. The utility of including INN in our formalism is that it allows generalization to both infinite systems where UE is not defined, and to real-world systems where UE is not physically observable (since, for example, $t_P$ could in principle be longer than the age of the universe). For the latter, INN can be an approximation to UE, where higher values of INN indicate a system more likely to exhibit UE (see Fig. \ref{INNREC}). Additionally, the combination of UE and INN can be used to exclude cases that appear unbounded but are only trivially so. For example, a partition of a system evolved according to a fixed dynamical rule could in principle locally satisfy UE, but would not satisfy INN since its dynamics could be shown to be equivalent to those generated from an appropriately constructed isolated system ({\it e.g.} a larger ECA in our example). An example is the time evolution of ECA Rule 30 \cite{NKS}, which is known to be a `complex' ECA rule that continually generates novel patterns under open-boundary conditions. In cases such as this, it should be considered that it is the complexity at the open boundary of the system that is generating continual novelty and not a mechanism {\it internal} to the system itself. In other words, in such examples the complexity is generated by the boundary conditions. Since our biosphere has simple, relatively homogeneous boundary conditions (geochemical and radiative energy sources) the complexity of the biosphere likely arises due to internal mechanisms and is not trivially generated by the boundary conditions alone \cite{smith2008thermodynamics}. Since we aim to understand the {\it intrinsic} mechanisms that might drive OEE in real, finite dynamical systems, we therefore require both definitions to be satisfied for a dynamical system to exhibit non-trivial OEE. 

\subsection*{Model implementation} \label{sec:MI}

We evaluate different mechanisms for generating OEE against Definitions \ref{Def:UE} and \ref{Def:INN}, utilizing the rule space of Elementary Cellular Automata (ECA) as a case study. ECA are defined as nearest-neighbor 1-dimensional CA operating on the two-bit alphabet $\{0, 1\}$. There are $256$ possible ECA rules, and since the rule numbering is arbitrary, we label them according to Wolfram's heuristic designation \cite{NKS}. Due to their relative simplicity, ECA represent some of the most widely-studied CA, thus providing a well-characterized foundation for this study. Traditionally, ECA evolve according to a fixed dynamical rule starting from a specified initial state. As such, no isolated finite ECA can meet both of the criteria laid out in Definitions \ref{Def:UE} and \ref{Def:INN} as per our construction aimed at excluding trivial cases. An isolated ECA of width $w$ will repeat its pattern of states by the Poincar\'e time $t_P = 2^{w}$ (violating Definition \ref{Def:UE}). If we instead considered a CA of width $w$ as a subsystem of a larger ECA it would not necessarily repeat within $2^{w}$ time steps, but it would {\it not} be innovative (violating Definition \ref{Def:INN}). Thus, as stated, we can exclude trivial examples such as ECA Rule 30, or other unbounded but non-innovative dynamical processes, which repeatedly apply the same update rule. A list of model parameters are summarized in Table \ref{Tab:parameters}.

\begin{table}[h]
\centering
\caption{Table of terms and model parameters.}
\begin{tabular}{cl}
                  &                                            \\
\textbf{Parameter} & \textbf{Definition}                              \\ \hline
                  &                                            \\
$o$               & Single organism execution                  \\
$e$               & Single environment execution               \\
$s_o$               & state of $o$                      \\
$r_o$               & rule of $o$                           \\
$s_e$               & state of $e$                                  \\
$w_o$               & width of $o$                          \\
$w_e$               & width of $e$   \\
$t_P$               & Poincar\'{e} recurrence time                                  \\
$t_r$			  & Recurrence time of $s_o$	\\
$t_r'$			  & Recurrence time of $r_o$							\\	
$I$			  & Innovation calculated as the normalized number of rule transitions						\\

$\mu$			  & Mutation threshold of Case III variant	$\mu = [0,1)$							\\
$\xi$			  & random noise for Case III variant, $\xi = [0,1)$		\\
$C$               & Compressibility                            \\
$k$               & Lyapunov exponent                         
\end{tabular} \label{Tab:parameters}
\end{table}

To exclude trivial unbounded cases, Definitions \ref{Def:UE} and \ref{Def:INN} are constructed to require that the dynamical rules themselves evolve in time. As we will show, utilizing the set of $256$ possible ECA rules as the rule space for CA with {\it time-dependent} rules makes both UE and INN possible. Rules can be stochastically or deterministically evolved, and we explore both mechanisms here. We note that there exists a huge number of possible variants one might consider. We therefore focus on three variants that display important mechanisms implicated in generating OEE, including openness to an environment \cite{ruiz2008enabling} (of varying degrees in all three variants),  state-dependent dynamics (regarded as a hallmark feature of life \cite{alglife, goldenfeld2011life, douglas1979godel}),  and stochasticity. Here openness to an environment is parameterized by the degree to which the rule evolution of $o$ depends on the state (or rule) of $o$, as compared to its dependence on the state of $e$. Completely open systems are regarded as depending only on external factors, such that the time-dependence of the rule evolution is {\it only} a function of the environment. We also consider cases that are only partially open, where the rule evolution depends on both extrinsic and intrinsic factors.

\subsubsection*{Case I}  

\begin{figure}
\centering
        \includegraphics[width=\linewidth]{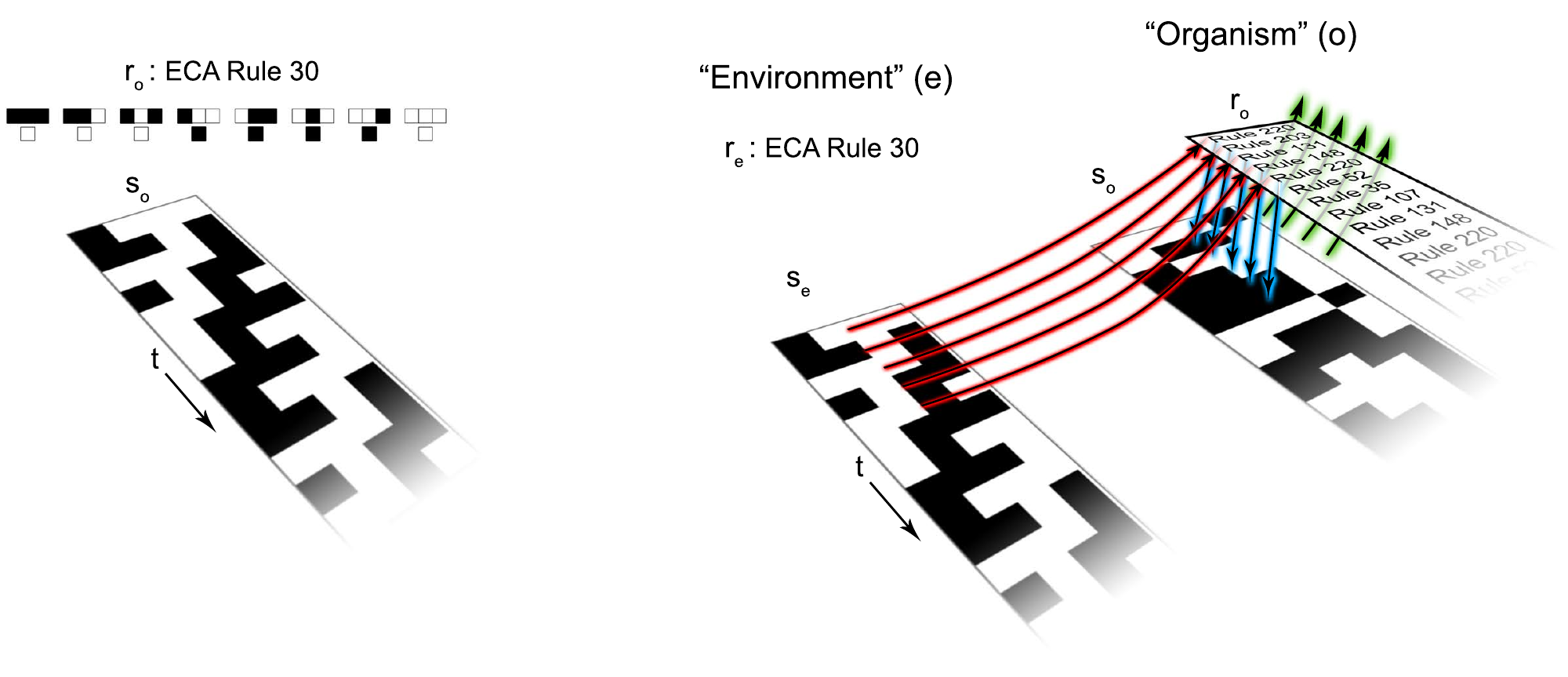}
  \caption{Illustrations of the time evolution of a standard ECA (left) and of a Case I state-dependent CA (right). ECA evolve according to a fixed update rule (here Rule 30), with the same rule implemented at each time step. In an ECA rule table, the cell representation of all possible binary ordered triplets is shown in the top row, with the cell representation of the corresponding mapping arising from Rule 30 shown below. Rule 30 therefore has the binary representation $00011110$. In a Case I CA (right), the environment subsystem $e$ evolves exactly like an ECA with a fixed rule. The organism subsystem $o$, by contrast, updates its rule at each time-step depending on its rule at the previous time-step, its own state (green arrows) and the state of $e$ (red arrows). The new rule for $o$ is then implemented to update the state of $o$ (blue arrows). The rules are therefore time-dependent in a manner that is a function of the states of $o$ and $e$ and the past history of $o$ (through the dependence on the rule at the previous time-step).} \label{fig:3DCAs}
\end{figure}

The first variant, Case I, implements {\it state-dependent} update rules, such that the evolution of $o$ depends on its own state {\it and} that of its environment. This is intended to provide a model that captures the hypothesized self-referential dynamics underlying biological systems (see {\it e.g.} Goldenfeld and Woese \cite{goldenfeld2011life}) while also being open to an environment (we do not consider closed self-referential systems herein as treated in Pavlic {\it et al} since these do not permit the possibility of UE \cite{pavlic2014self}). We consider two coupled subsystems $o$ and $e$, where the update rule of $o$ is state-dependent and is a function the state and rule of $o$, and the state of $e$ at the same time $t$ (thus being self-referential but also open to perturbations from an external system). That is, the update rule of $o$ takes on the functional form $r_o(t+1) = f(s_o(t), r_o(t), s_e(t))$, where $s_o$ and $r_o$ are the state and rule of the organism respectively, and $s_e$ is the state of the environment. We regard this case as only partly open to an environment since the evolution of the rule of $o$ depends on its own state (and rule) in addition to the state of its environment. By contrast, the subsystem $e$ is closed to external perturbation and evolves according to a {\it fixed} rule (such that $e$ is an ECA ). Both $o$ and $e$ have periodic boundary conditions ($o$ is only open in the sense that its rule evolution is in part externally driven). A schematic illustration of the time evolution of an ECA, and the coupling between subsystems in a Case I CA is shown in Figure \ref{fig:3DCAs}. 

\begin{figure}
    \includegraphics[width=\linewidth]{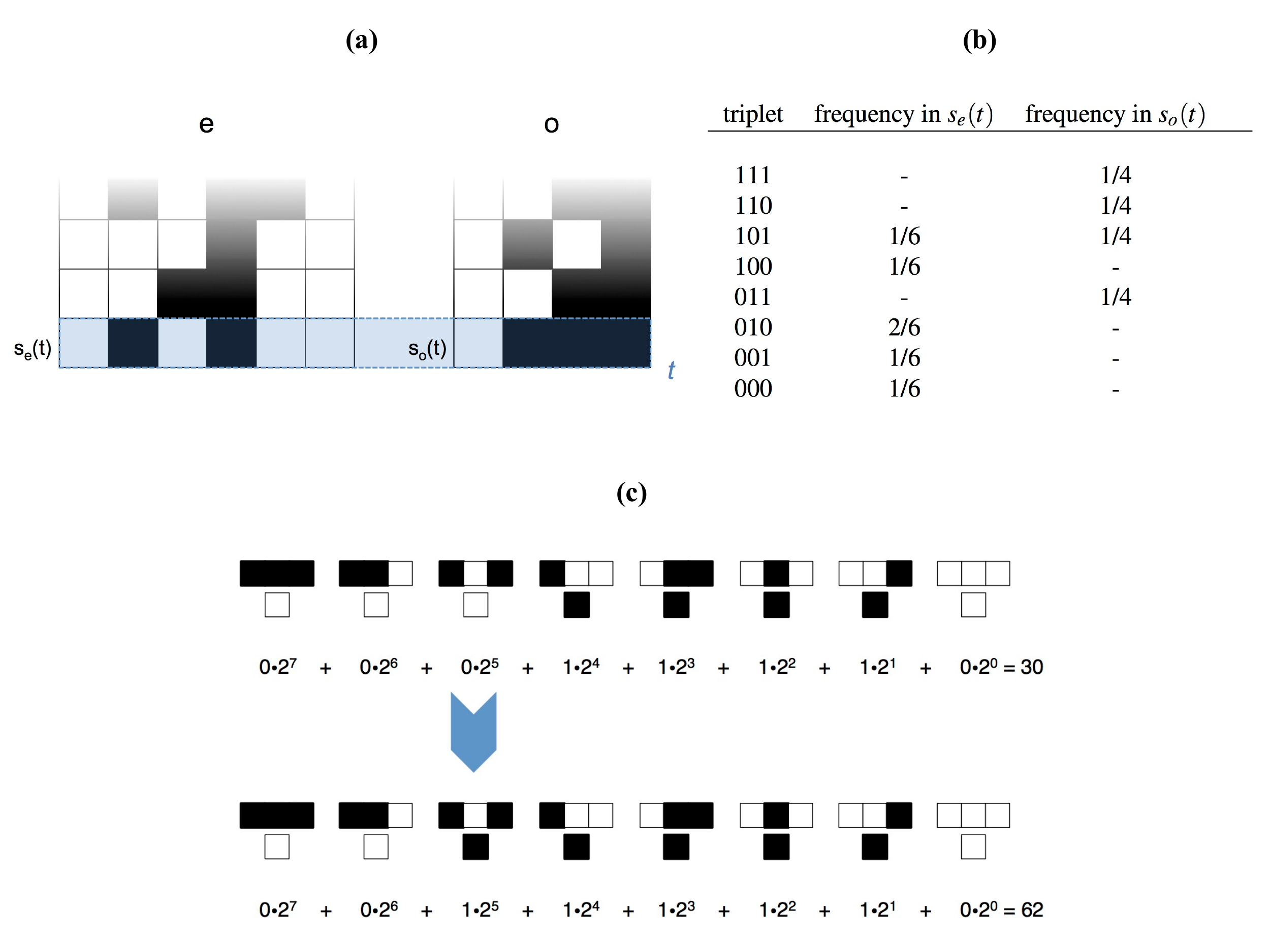}
\caption{Example of the implementation of a Case I organism in our example. Shown is an organism $o$ of width $w_o = 4$, coupled to an environment $e$ with width $w_e = 6$, where the rule of $o$ at time step $t$ is $r_o(t) = 30$.  {\bf (a)} At each time step $t$, the frequency of ordered triplets are compared in the state of the organism and that of the environment, $s_o$ and $s_e$ respectively, and used to update $r_o(t) \rightarrow r_o(t+1)$ (see text for algorithm description). {\bf (b)} Table of the calculated frequency of ordered triplets in the state of the environment and in the state of the organism for time step $t$ shown in the left panel. {\bf (c)} Update of $r_o$ from Rule 30 to Rule 62, based on the frequency of triplets in the table \textbf{(b)}.} \label{fig:compare}
\label{fig:test}
\end{figure}

To demonstrate how the organism in our example of a Case I CA changes its update rule, we provide a simple illustrative example of the particular function $f(s_o(t), r_o(t), s_e(t))$ implemented in this work (see \ref{fig:compare} and Supplement 1.1). Specifically, we utilize an update function that takes advantage of the binary representation of ECA. An example of the structure of an ECA rule is shown for Rule 30 in Fig. \ref{fig:3DCAs}.  ECA rules are structured such that each successive bit in the binary representation of the rule is the output of one of the $2^3$ possible ordered sets of triplet states. The left panel of Fig. \ref{fig:compare} shows an example of a few times steps of the evolution of an organism $o$ of width $w_o = 4$ (right) coupled to an environment $e$ with width $w_e = 6$ (left), where $o$ implemented rule 30 at $t-1$. At each time-step $t$ the frequency of each of the $2^3$ ordered triplet states (listed in the top row of Fig. \ref{fig:3DCAs}) in the state of $o$ is compared to the frequency of the same ordered triplet in the state of $e$. If the frequency in $o$ meets or exceeds the frequency in $e$ for a given triplet, the bit corresponding to the output of that triplet in the rule of $o$ is flipped from $0 \leftrightarrow 1$. For the example in the left panel of Fig. \ref{fig:compare}, the triplet frequencies are listed in the table in the right panel of Fig. \ref{fig:compare}. We note that for our implementation, the frequency of a triplet in $o$ is calculated relative to the total number of possible triplets in $s_o$, which is $4$ in this example (and likewise for $e$, with $6$ possible triplets in the current state). We compare the frequency only for those triplets that appear in the state of $o$ at time $t$. In the table, only the triplet $101$ is expressed more frequently in the organism $o$ than in the environment $e$.  The interaction between $o$ and $e$ changes $r_o$ from Rule 30 at time-step $t$ to Rule 62 at $t+1$, as shown in Fig. \ref{fig:compare}.  The rule may change by more than one bit in its binary representation in a single time step if multiple triplets meet the criteria to change the organism's update rule.

We note that we do not expect the qualitative features of Case I CA reported here to depend on the precise form of the state-dependent update rule as presented, so long as the update of $r_o$ depends on the state and rule of $o$, and the state of $e$ (that is, $o$ is self-referential {\it and} open -- see Pavlic {\it et al.} \cite{pavlic2014self} for an example of non-open self-referencing CA that does not display UE). We explored some variants to this rule-changing mechanism. None of our variants significantly changed the statistics of the results, indicating that the qualitative features of the dynamics do not depend on the exact (and somewhat parochial) details of the example presented herein. Instead, we regard the important part to be the general feature of self-reference coupled to openness to an environment that is driving the interesting features of the dynamics observed, where we can focus on just one example in this study for computational tractability in generating large ensemble statistics. The example implemented here was chosen since it takes explicit advantage of the structure of ECA rules (by flipping bits in the rule table) to provide a simple, open state-dependent mechanism for producing interesting dynamics. 

\subsubsection*{Case II}

We introduce a second variant of CA, Case II, that is similarly composed of two spatially segregated, fixed-width, 1-dimensional CA: an organism $o$ and an environment $e$. As with Case I, the environment $e$ is an execution of an ECA, and is evolved according to a fixed rule drawn from the set of $256$ possible ECA rules. The key difference between Case I and Case II CA is that for Case II, the the update rule of the subsystem $o$ depends {\it only} on the state of the external environment $e$ and is therefore independent of the current state or rule of $o$ -- that is, $o$ is {\it not} self-referential in this example. Case II CA emulate systems where the rules for dynamical evolution are modulated exclusively by the time evolution of an external system. We consider $o$ in this example to be more open to its environment than for Case I, since the rule evolution of $o$ depends only on $e$. The functional form of Case II rule evolution may be written as $r_o (t + 1) = f (s_e (t ))$, where $r_o$ is the rule of $o$ and $s_e$ is the state of the environment (see Supplement 1.2). For the example presented here, we implement a map $f$ that takes $s_e(t) \rightarrow r_o(t)$ that is $1:1$ from the state of $e$ to the binary representation of the rule of $o$ (determined according to Wolfram's binary classification scheme). Therefore for the implementation of Case II in our example the environment must be of width $w_e = 8$. 

\subsubsection*{Case III}

The final variant, Case III, is composed of a single, fixed-width, 1-dimensional CA with periodic boundary conditions, which is identified as the organism $o$. Like with Case II, the rule evolution of Case III is driven externally and does not depend on $o$. However, here the external environment $e$ is stochastic noise and not an ECA.  The subsystem $o$ has a time-dependent rule where each bit in the rule table is flipped with a probability $\mu$ (``mutation rate'') at each time step. In functional form, the subsystem $o$ updates its rule such that  $r_o (t + 1) = f (r_o (t ), \xi )$, where $r_o$ is the rule of $o$, and $\xi$ is a random number drawn from the interval $[0,1)$ (see Supplement 1.3). At each time step, for each bit in the rule table, a random number $\xi$ is drawn, and if $\xi$ is above a threshold $\mu$, that bit is flipped $0 \leftrightarrow 1$ at that time step. This implements a diffusive-random walk through ECA rule space. Since the rule of $o$ at time $t + 1$, $r_o (t + 1)$, depends on the rule at time $t$, $r_o(t)$, the dynamics of Case III CA are history-dependent in a similar manner to Case I (both rely on flipping bits in $r_o(t)$, where Case I do so deterministically as a function of $s_o$ and $s_e$, and Case III do so stochastically).  In this example, $o$ is also more open to its environment than in Case I since the organism's rule does not depend on $s_o$, but it is less open than Case II since the rule does depend on the previous organism rule used.

\begin{table}[h!]
\centering
\caption{Table of cellular automata variants, and the functional form of the rule evolution of $o$.}
\label{modeltable}
\begin{adjustbox}{max width=0.6\textwidth}
\begin{tabular}{| l | l | l |}     
\hline
\textbf{CA Variant}& \textbf{Organism Rule Evolution} & \textbf{Environment, $e$}\\ \hline
\multicolumn{1}{|l|}{\begin{tabular}[c]{@{}l@{}}Case I  \end{tabular}}   & \multicolumn{1}{l|}{\begin{tabular}[c]{@{}l@{}} $r_o(t+1) = f (s_o(t), r_o(t), s_e(t))$\end{tabular}} & \multicolumn{1}{l|}{\begin{tabular}[c]{@{}l@{}}ECA, varied $w_e$\end{tabular}} \\ \hline
\multicolumn{1}{|l|}{\begin{tabular}[c]{@{}l@{}}Case II \end{tabular}} & \multicolumn{1}{l|}{\begin{tabular}[c]{@{}l@{}} $r_o(t + 1) = f(s_e(t))$\end{tabular}}                               & \multicolumn{1}{l|}{\begin{tabular}[c]{@{}l@{}}ECA, $w_e=8$\end{tabular}}     \\ \hline
\multicolumn{1}{|l|}{\begin{tabular}[c]{@{}l@{}}Case III \end{tabular}}       & \multicolumn{1}{l|}{\begin{tabular}[c]{@{}l@{}} $r_o(t + 1) = f(r_o(t), \xi)$ \end{tabular}}                                         & \multicolumn{1}{l|}{Heat bath}        \\ \hline
\multicolumn{1}{|l|}{ECA (Isolated)}         & \multicolumn{1}{l|}{\begin{tabular}[c]{@{}l@{}}$r_o(t + 1) = r_o(t)$ \end{tabular}}                                           & \multicolumn{1}{l|}{None}             \\ \hline
\end{tabular} \label{tab:CAs}
\end{adjustbox}
\end{table}

All three variants are summarized in Table \ref{tab:CAs} (see Supplement 1), where the functional dependencies of the rule evolution in each example are explicitly compared. Since we restrict the rule space for Cases I--III to that of ECA rules only, the trajectories of ECA with periodic boundary conditions provides a well-defined set of isolated counterfactual trajectories with which to evaluate Definitions \ref{Def:UE} and \ref{Def:INN}. For comparison to isolated systems, we evaluate {\it all} ECA of width $w_o$, where $w_o$ is the width of the ``organism'' subsystem $o$. We test the capacity for each of the three cases presented to generate OEE against Definitions \ref{Def:UE} and \ref{Def:INN} in a statistically rigorous manner, and compare the efficacy of the different mechanisms implemented in each case. 

\subsubsection*{Experimental Methods}
For Cases I - III, we evolve $o$ with periodic boundary conditions (such that interaction with the environment is only through the rule evolution). For Cases I and II, $e$ is also a CA with periodic boundary conditions. For Case I, where $w_e$ must also be specified, we consider systems with $w_e = 1/2 w_o$, $w_o$, $3/2 w_o$, $2w_o$ and $5/2 w_o$ , where $w_o$ is the width $o$. For Case II, $w_e = 8$ for all simulations, since this permits a 1:1 map from the possible states of $e$ to the rule space of ECA. Results for Case III are given for organism rule mutation rate $\mu = 0.5$, such that each outcome bit in the rule evolution has a 50\% probability of flipping at every time step for $\xi$ drawn from the interval $[0,1)$ (a bit flips when $\mu > \xi$). Other values of $\mu$ were explored, with qualitatively similar results (see Supplement Fig. S4). 

The number of possible executions grows exponentially large with width $w_o$, limiting the computational tractability of statistically rigorous sampling. We therefore explored small CA with $w_o = 3, 4,\ldots 7$ and sampled a representative subspace of each (see Supplement 2). For each system sampled, we measured the recurrence times of the rule ($t_r'$) and state ($t_r$) trajectories for $o$. For Case III CA, which are stochastic, all simulations eventually terminated as a random oscillation between the all $`0`$ state and the all $`1`$ state. We therefore used the timescale of reaching this oscillatory attractor as a proxy for the state recurrence time $t_r$. In cases where $t_r > t_P$ or $t_r' > t_P$, where $t_P = 2^w_o$ for isolated ECA (Definition \ref{Def:UE}), and the state trajectory was not produced by {\it any} ECA execution of width $w_o$ (Definition \ref{Def:INN}), the system is considered to exhibit OEE.

We measured the complexity of the resulting interactions by calculating relative compressibility, $C$, and by the system's sensitivity based upon Lyapunov exponents, $k$~\cite{lyp} (see Supplement 8). Large values of $C$ indicate low Kolmogorov-Chaitin complexity, meaning the output can be produced by a simple (short) program.  Large values of $k$ indicate complex dynamics, with trajectories that rapidly diverge for small perturbations such as occurs in deterministic chaos \cite{lyp}. These values are compared to those of ECA. Additionally, ECA rules are often categorized in terms of four Wolfram complexity classes, I - IV \cite{NKS}. Class I and II are considered simple because all initial patterns evolve quickly into a stable or oscillating, homogeneous state. Class III and IV rules are viewed as generating more complex dynamics. We use the complexity classes of the rules utilized in time-dependent rule evolution to determine whether the complexity of time-dependent CAs is a product of the ECA rules implemented, or if it is generated through the mechanism of time-dependence.

\section*{Results}

The vast majority of executions sampled from all three CA variants were innovative by Definition \ref{Def:INN}, with $> 99\%$ of Case II and Case III CAs displaying INN. For Case I CA, the percentage of INN cases increased as a function of both  $w_o$ and $w_e$, ranging from $\sim 30\%$ for the smallest CA explored to $>99\%$ for larger systems (see Supplement 5). This is intuitive, since the majority of organisms with changing updates rules should be expected to exhibit different state-trajectories than ECA. The fact that $>99\%$ of organisms are innovative in our examples may seem to indicate that INN is trivial. However, we note that INN conceptually becomes more significant when considering infinite systems (where UE is not defined) or large systems where $t_P$ is not measurable (and thus UE cannot be calculated). We show below that INN scales with recurrence time, and the amount of innovation is a good proxy for UE. INN is therefore useful to the analysis of large or infinite systems where the methods implemented here to detail candidate mechanisms are not directly applicable to test UE. INN is also necessary to exclude trivial OEE. We also note that for computational tractability we compare the time evolution of $o$ only to ECA, but in practice one could (and perhaps should) compare $o$ to dynamical systems evolved according to {\it any} fixed rule ({\it e.g.} regardless of neighborhood size, which for ECA is $n = 3$), in which case we might expect the number of INN cases to decrease and therefore INN would be more non-trivial even for small systems. 

By contrast to cases exhibiting INN, OEE cases are much rarer, even for our highly simplified examples, due to the fact that the number of UE cases is much smaller, typically representing $<5\%$ of all the sampled trajectories in the examples studied here. We therefore focus discussion primarily on sampled executions meeting the criteria for OEE, {\it i.e.} those that satisfied Definitions \ref{Def:UE}, before returning to how INN might approximate UE.

\begin{table}
\centering
\caption{Percentage of sampled cases displaying OEE (satisfying Definitions \ref{Def:UE} and \ref{Def:INN}) for each CA variant. Rows are organism width, $w_o$, and columns correspond to the three different CA variants and ECA statistics. } \label{tab:OEE_percent}
\begin{tabular}{l|llll}
{$w_o$}& \textbf{ECA} & \textbf{Case I ($w_o = w_e$)} & \textbf{Case II} & \textbf{Case III} \\ \hline
\textbf{3} & $0$ & $0.02$ & $42.47$ & $7.42$ \\ 
\textbf{4} & $0$ & $0.38$ & $11.54$ & $1.05$  \\ 
\textbf{5} & $0$ & $3.41$ & $10.43$ & $2.76$  \\ 
\textbf{6} & $0$ & $0.03$ & $0.27$ & $5.2\times 10^{-3}$ \\ 
\textbf{7} & $0$ & $1.06$ & $0.7$ & $4.7\times 10^{-4}$  \\ 
\end{tabular}
\end{table}

\begin{figure}
    \centering
\includegraphics[width=0.5\textwidth]
    {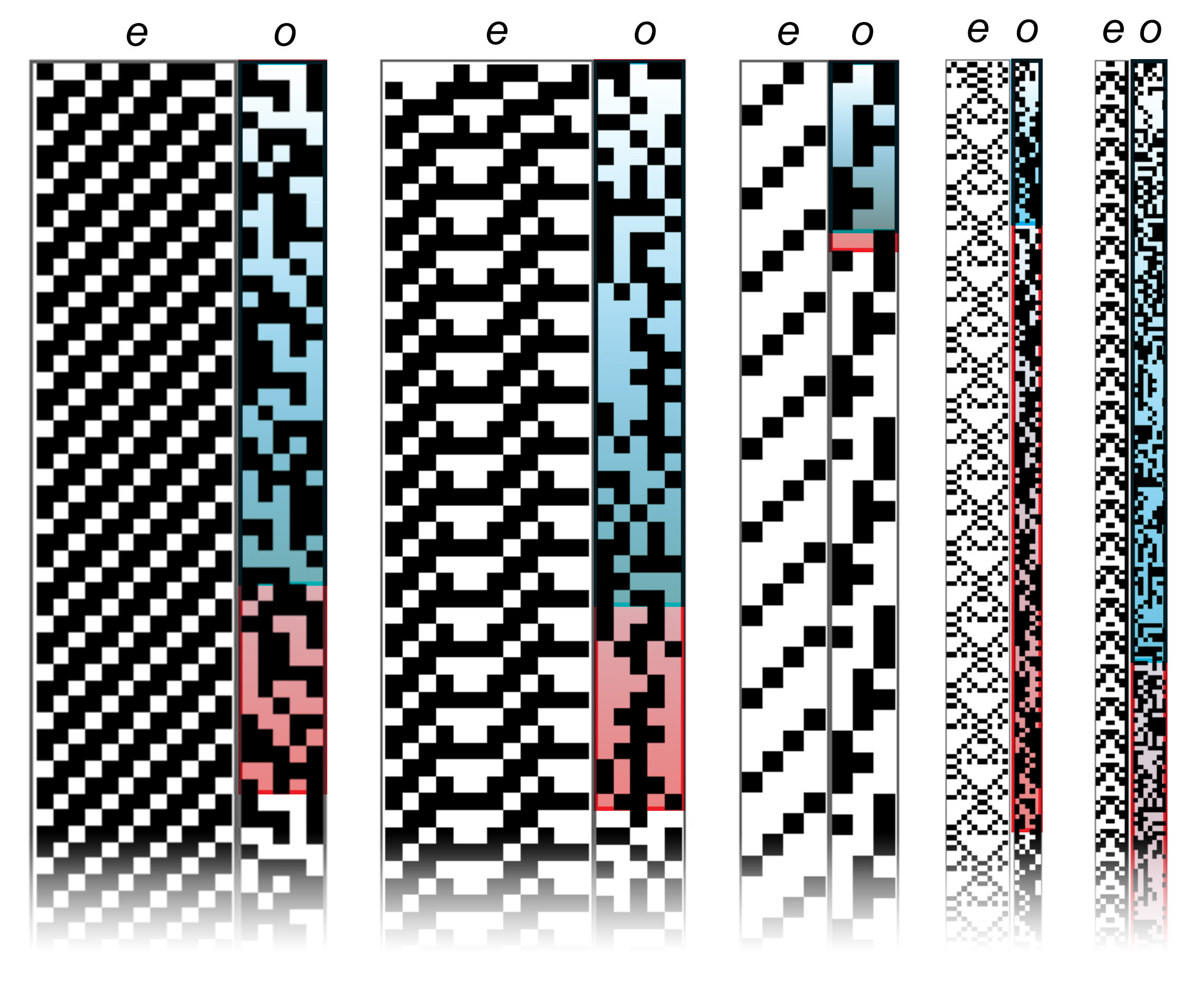}
    \caption{Examples of Case I CA exhibiting OEE. In each panel the environment $e$ is shown on the left, and organism $o$ on the right. For each $o$, the Poincar\'e recurrence time ($t_P$) for an isolated system is highlighted in blue, and the recurrence time of the states of $o$, $t_r$, is highlighted in red.}
    \label{fig:eco_CAs}
\end{figure}

\subsection*{Open-Ended Evolution in CA variants.}

\begin{figure}
        \centering
        \includegraphics[width=\linewidth]{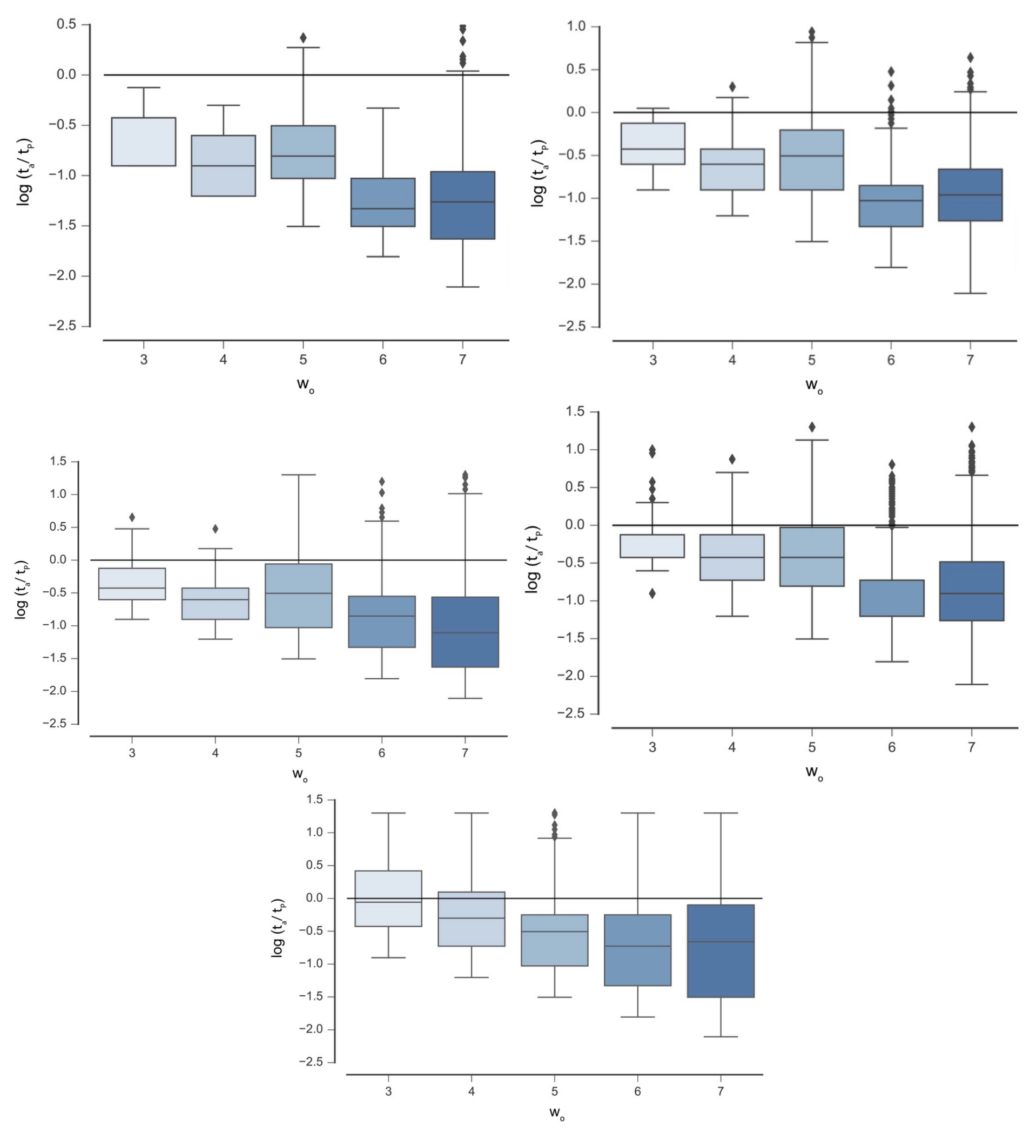}
    \caption{Distribution of recurrence times $t_r$ for the state trajectory of $o$ for Case I CA. From top to bottom are distributions for $w_e = 1 w_o, w_o, 3 w_o, 2w_o$ and $5 w_o$, respectively. In all panels the black horizontal line indicates where $tr/tP = 1$ (shown on a log scale). Sampled trajectories displaying UE occur for $t_r/t_P > 1$.}
    \label{fig:tr1}
\end{figure}
\begin{figure}
        \centering
        \includegraphics[width=\linewidth]{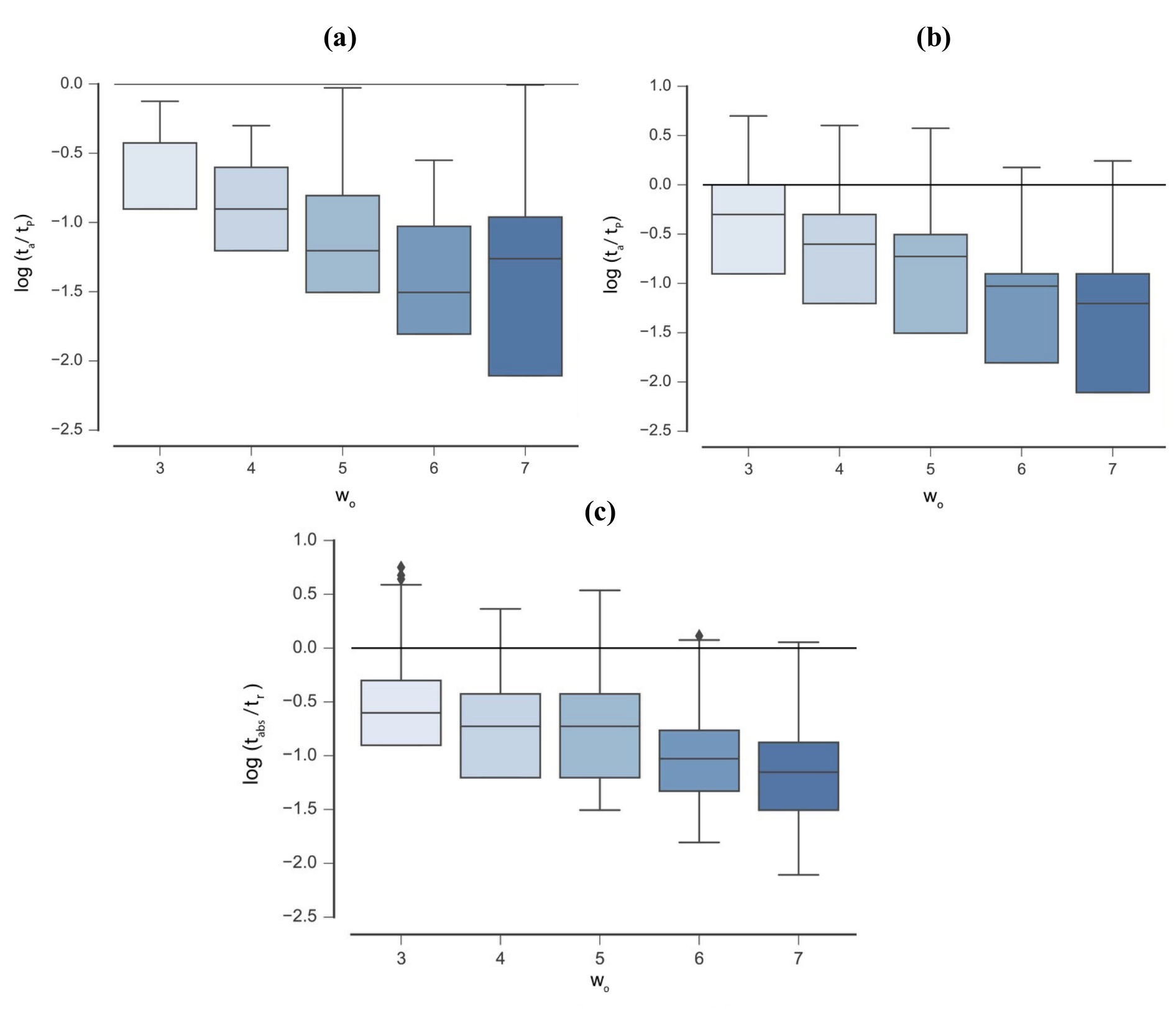}
    \caption{Distribution of recurrence times $t_r$ for the state trajectory of $o$ for ECA (top left), Case II (top right), and Case III CA (bottom). In all panels the black horizontal line indicates where $tr/tP = 1$ (shown on a log scale). Sampled trajectories displaying UE occur for $t_r/t_P > 1$.}
    \label{fig:tr2}
\end{figure}

The percentage of sampled cases for each CA variant that satisfy Definitions \ref{Def:UE} for UE are shown in Table \ref{tab:OEE_percent}, where for purposes of more direct comparison Case I CA statistics are shown only for $w_o = w_e$. Case I CA statistics for other relative values of $w_o$ and $w_e$ are shown in Table \ref{tab:oeetable_variants2}. Box plots of the distribution of measured recurrence times for each CA variant are shown in Fig. \ref{fig:tr1} and Fig. \ref{fig:tr2}. All UE cases presented here are also INN, and thus exhibit OEE. We therefore refer to UE and OEE interchangeably (without explicitly referencing OEE as cases exhibiting UE and INN separately). Examples of Case I CA exhibiting OEE are shown in Fig. \ref{fig:eco_CAs}, demonstrating the innovative patterns that can emerge due to time dependent rules. 

To compare the capacity for OEE across the different CA variants tested, it is useful to define a notion of scalability \cite{taylor2016open}. Here we define {\bf scalable} systems as ones where the number of observed OEE cases can increase without the need to either (1) change the rule-updating mechanism of $o$ or (2) significantly change the statistics of sampled cases. By this definition, the two primary mechanisms for increasing the number of OEE cases in a scalable manner are by changing $w_o$, or depending on the nature of the coupling between $o$ and $e$, changing $w_e$ (with the constraint that the rule-updating mechanism cannot change).  

As expected (by definition), isolated ECA do not exhibit any OEE cases and the majority of ECA have recurrence times $t_r \ll t_p$. However, all three CA variants with time-dependent rules do exhibit examples of OEE, but differ in the percentage of sampled cases and their scalability. Case III exhibits the simplest dynamics, where trajectories follow a diffusive random walk through rule space until the system converges on a random oscillation between the all- $'0'$ and all-$'1'$ states (where $t_r$ is approximated by this convergence time). The frequency of state recurrence times $t_r$ of the organism decreases exponentially (see Supplement Fig. S4), such that the $o$ with the longest recurrence times are exponentially rare. Since so few examples were found for organisms of size $w_o = 7$, we also tested $w_o = 8$ and found no examples of OEE. In general, the exponential decline observed is steeper for increasing $w_o$. Observing more OEE cases therefore requires exponentially increasing the number of sampled trajectories for increasing $w_o$. The capacity for Case III CA to demonstrate OEE is therefore not scalable with system size (violating condition 2) in our definition above). An additional limitation of Case III CA is that their the long-term dynamics are relatively simple once the system settles into the oscillatory attractor, thus the majority of observations of Case III CA would not yield interesting dynamics ({\it e.g.} if the observation time were much greater than the start time $t_{obs} \gg t_o$).

 \begin{table}
\centering
\caption{Percentage of sampled cases displaying OEE (satisfying Definitions \ref{Def:UE} and \ref{Def:INN}) for Case I, with varying environment size $w_e$.} 
\begin{tabular}{l|lllll}
{$w_o$}& \textbf{$\frac{1}{2} w_o$} & \textbf{$w_o$} & \textbf{$\frac{3}{2} w_o$} & \textbf{$2 w_o$} & \textbf{$\frac{5}{2} w_o$}  \\ \hline
\textbf{3} & $0$ & $0.02$ & $6.52$ & $10.81$ & $28.14$  \\ 
\textbf{4} & $0$ & $0.38$ & $2.28$ & $2.94$ & $9.65$  \\ 
\textbf{5} & $0$ & $3.41$ & $7.04$ & $7.5$ & $8.64$  \\ 
\textbf{6} & $0$ & $0.03$ & $2.15$ & $2.64$ & $5.82$  \\ 
\textbf{7} & $0$ & $1.06$ & $2.95$ & $4.39$ & $5.34$   \\ 
\end{tabular}
\label{tab:oeetable_variants2}
\end{table}

For Case II, we also observed a steep decline in the number of OEE cases observed for increasing $w_o$ (Table \ref{tab:OEE_percent}). This is reflected by a steady decrease in the mean of the recurrence times for increasing $w_o$, as shown in Fig. \ref{fig:tr2}. We also tested a large statistical sample of organisms of size $w_o \geq 8$ for Case II CA (not shown) and found no examples of OEE cases. This is not wholly unexpected. For Case II with $w_o = 8$, the environment and organism are the same size ($w_e = w_o$). Therefore $e$ and $o$ share the same Poincar\'e time $t_P = 2^{w_o}$. The subsystem $e$ is a traditional ECA, therefore the majority of $e$ will exhibit recurrence times $\ll t_P$ (see {\it e.g.} trend in Fig. \ref{fig:tr2}). Since the rule of $o$ is determined by a 1:1 map from the state of $e$, the rule recurrence time of $o$ will also be much less than the Poincar\'e time, such that $t_r' \ll t_P$. It is the rule evolution that drives novelty in the state evolution, we therefore also see that the state recurrence time is similarly limited such that $t_r \ll t_P$ also holds. To get around this limitation one could increase the size of the environment such that $w_e > w_o$. However, since the rule for $o$ is a 1:1 map from the state of $e$, this would require changing the updating rule scheme for $o$. That is, the organism $o$ would have to change how it evolves in time as a function of its environment (violating 1) in our definition of scalability above. By our definition of scalability, this is not a scalable mechanism for generating OEE since $o$ must change the function for its updating rule and therefore would represent a different $o$. 

We can compare the statistics of sampled OEE cases for Case I where $w_o = w_e$ to those of Case II and Case III, as in Table \ref{tab:OEE_percent}. While Case II and Case III CA see a steep drop-off in the percentage of sampled cases exhibiting OEE with increasing organism size $w_o$, the Case I CA exhibit a flatter trend. We determined whether this trend holds for varying $w_e$ by also analyzing statistics for Case I CA where $w_e = \frac{1}{2}w_o$, $w_o$, $\frac{3}{2}w_o$, $2w_o$ and $\frac{5}{2}w_o$. The statistics of OEE cases sampled are shown in Table \ref{tab:oeetable_variants2} and box plots of the distribution of recurrence times are shown in Fig. \ref{fig:tr1}. For each fixed environment size explored ($w_e$, columns in Table \ref{tab:oeetable_variants2}), we observe that the statistics do not decrease dramatically as the size of the organism increases (increasing $w_o$). For fixed organism size ($w_o$, rows in Table \ref{tab:oeetable_variants2}), we observe that the number of OEE cases {\it increases} with increasing environment size. These trends are also reflected in the means of the distributions shown in Fig. \ref{fig:tr1}. Case I represents a scalable mechanism for OEE as $o$ can be coupled to larger environments and will produce more OEE cases. 

Case I and Case II can be contrasted to gain insights into scalability. The key difference between the two variants is that for Case II the update rule of $o$ is a 1:1 map with the state of $e$, whereas for Case I the map is self-referencing and is {\it many}:1. Case I therefore uses a {\it coarse-grained} representation of the environment for updating the rule of $o$ and because the dynamics are self-referential, the same pattern in the environment can lead to different rule transitions in $o$, depending on the previous state and rule of $o$. Thus, although both Case I and Case II exhibit trends of increasing OEE as $w_e$ is increased relative to $w_o$, the degree to which the size of the environment can impact the time evolution of the organism is different for the two cases. For a comparable size environment in Case I and Case II CA, the pattern relevant to the update of $o$ may have a longer recurrence time than the actual states of $e$ for Case I CA (due to the coarse-graining), whereas for Case II CA this pattern is strictly limited by the environment's recurrence time. Additionally, due to the coarse-graining of the environment in Case I CA, the update rule of $o$ is not dependent on the size of $e$: the same exact function for updating the rule of $o$ may be applied {\it independent} of the environment size. This is not true for Case II, where the function for updating the rule of $o$ must change in order to accommodate larger environments. 

\subsection*{INN as a proxy for UE}
\begin{figure}[ht!]
    \centering
        \includegraphics[width=\textwidth]{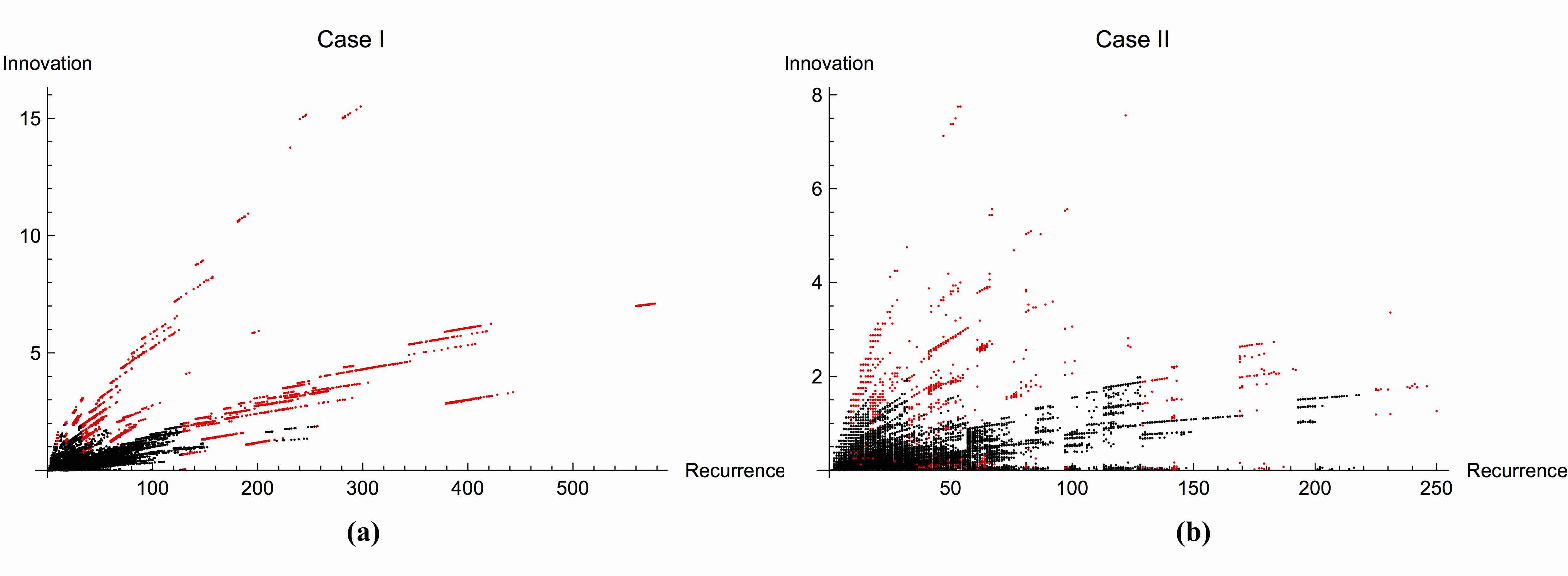} \label{fig:Inn_caseI}
    \caption{Relative innovation as a function of recurrence times for Case I (left) and Case II (right) CA. Highlighted in red are cases exhibiting OEE.}    \label{INNREC}
\end{figure}

We have presented examples of small dynamical systems to perform rigorous statistical testing of INN and UE to evaluate candidate mechanisms for generating OEE. An important question is how the results might apply to larger dynamical systems that could depend on different mechanisms than those testable in simple, discrete systems. While an approximation of INN is in principle measurable for large or infinite dynamical systems, UE is not measurable or not well-defined. We therefore aimed to determine if INN can be utilized as a proxy for UE. To do so, we defined a new parameter $n_r$, which quantifies the number of times that an organism changed its update rule between two successive time steps in its dynamical evolution. We normalized to determine the relative {\bf innovation} of an organism $I = \frac{n_r}{2^w}$ to generate a standardized measure for comparing across example organisms in our study. Statistically representative results for Case I and Case II organisms are shown in Fig. \ref{INNREC}, where $I$ is plotted against the organism's state recurrence time (Case III results are not included since the recurrence time is not well-defined).  For both Case I and Case II a clear trend is apparent where innovation is positively (and nearly linearly) correlated with recurrence time. For a given recurrence time, OEE cases (highlighted in red) are the most innovative. Comparing the two panels, it is evident that Case I CA exhibit higher innovation and therefore achieve longer recurrence times than Case II CA.  From these results we can conclude that a statistical measure sampling the number of observed rule transitions could be used as a proxy for UE, which we leave as a subject for future work. 

\subsection*{On-going Generation of Complexity in Case I}
\begin{figure*}
\centering
\includegraphics[width=\linewidth]{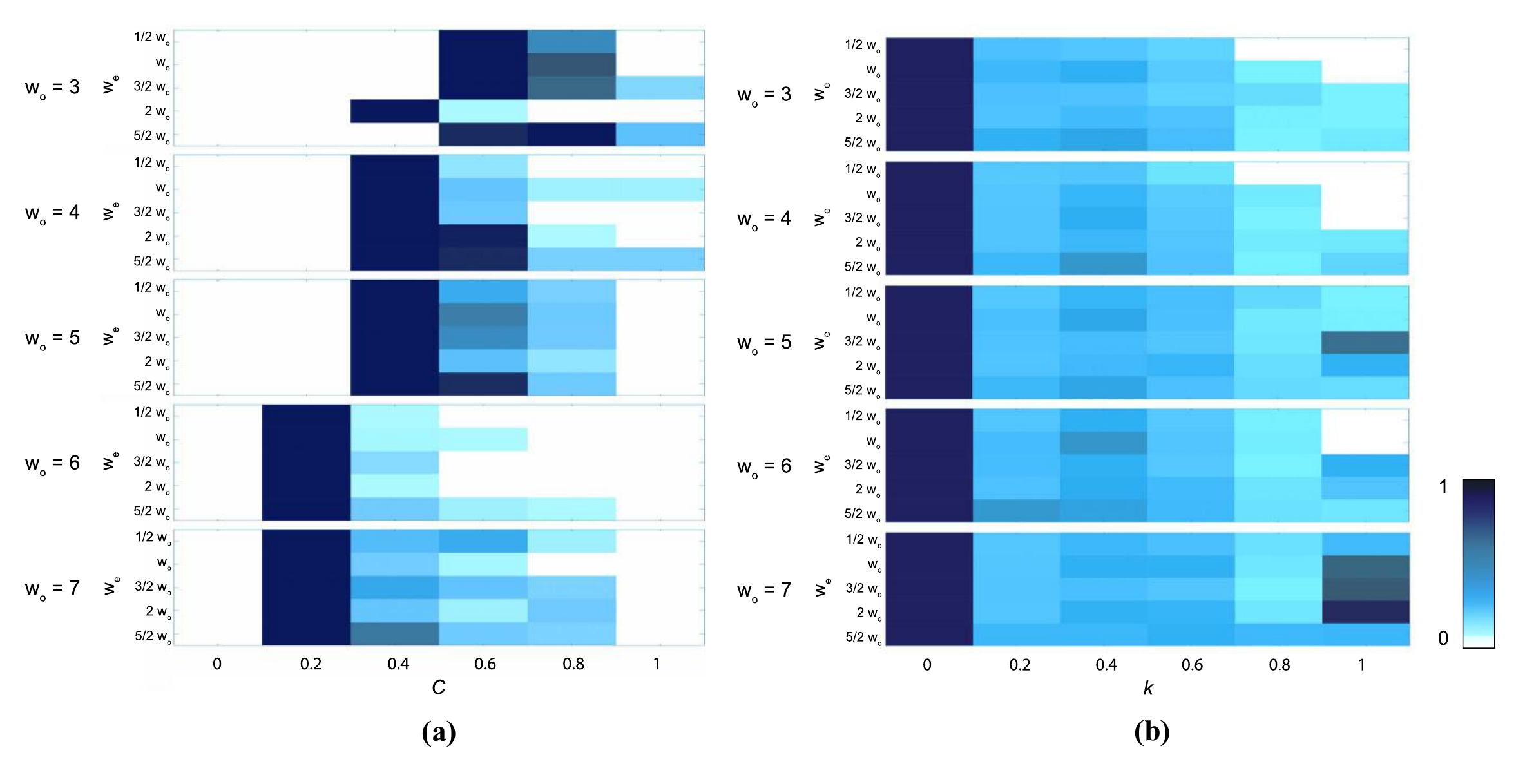}
\caption{Heat maps of compression $C$ (left) and Lyapunov exponent values $k$ (right) for all state trajectories of sampled $o$ for Case I CA. From top to bottom $w_o  = 3, 4, 5, 6$ and $7$, with distributions shown for $w_e = \frac{1}{2}w_o$, $w_o$, $\frac{3}{2}w_o$, $2w_o$ and $\frac{5}{2}w_o$ (from top to bottom in each panel, respectively) for each $w_o$. Distributions are normalized to the total size of sampled trajectories for each $w_o$ and $w_e$ (see statistics in Table S3).} \label{fig:heatmaps}
\end{figure*}

We also considered the complexity of Case I CA, relative to isolated ECA, as a further test of their scalability and potential to generate complex and novel dynamics. We characterized the complexity of Case I using two standard complexity measures, compressibility ($C$) and Lyapunov exponent ($k$). The trends demonstrate that in general $C$ decreases with increasing organisms width $w_o$, but increases with increasing environment size $w_e$ (left panel, Fig. \ref{fig:heatmaps}), indicative of increasing complexity with organism width $w_o$. Similar trends are observed for the Lyapunov exponent, as shown in the right panel of Fig.~\ref{fig:heatmaps}, where it is evident that increasing $w_o$ {\it or}  $w_e$ leads to an increasing number of cases with higher Lyapunov exponent $k$. OEE cases tend to have the highest $k$ values (see Supplement Fig. S12). As $C$ is normalized relative to ECA (see Supplement 8), we conclude that Case I CA are generally more complex than ECA evolved according to fixed dynamical rules, and this is especially true for OEE cases.

We also analyzed the ECA rules implemented in sampled Case I trajectories relative to the Wolfram Rule complexity classes. We find that Case I CA, on average, implement more Class I and II rules than Class III or IV, as shown in the frequency distribution of Fig. \ref{fig:metagenome} for Case I CA with $w_o = w_e$ (see Supplement Fig. 5 and 6).  Thus, we can conclude that the complexity generated by Case I CA is {\it intrinsic} to the state-dependent mechanism, and is not attributable to Class III and Class IV ECA rules dominating the rule evolution of $o$.

\begin{figure*}
\centering
\includegraphics[width=\linewidth]{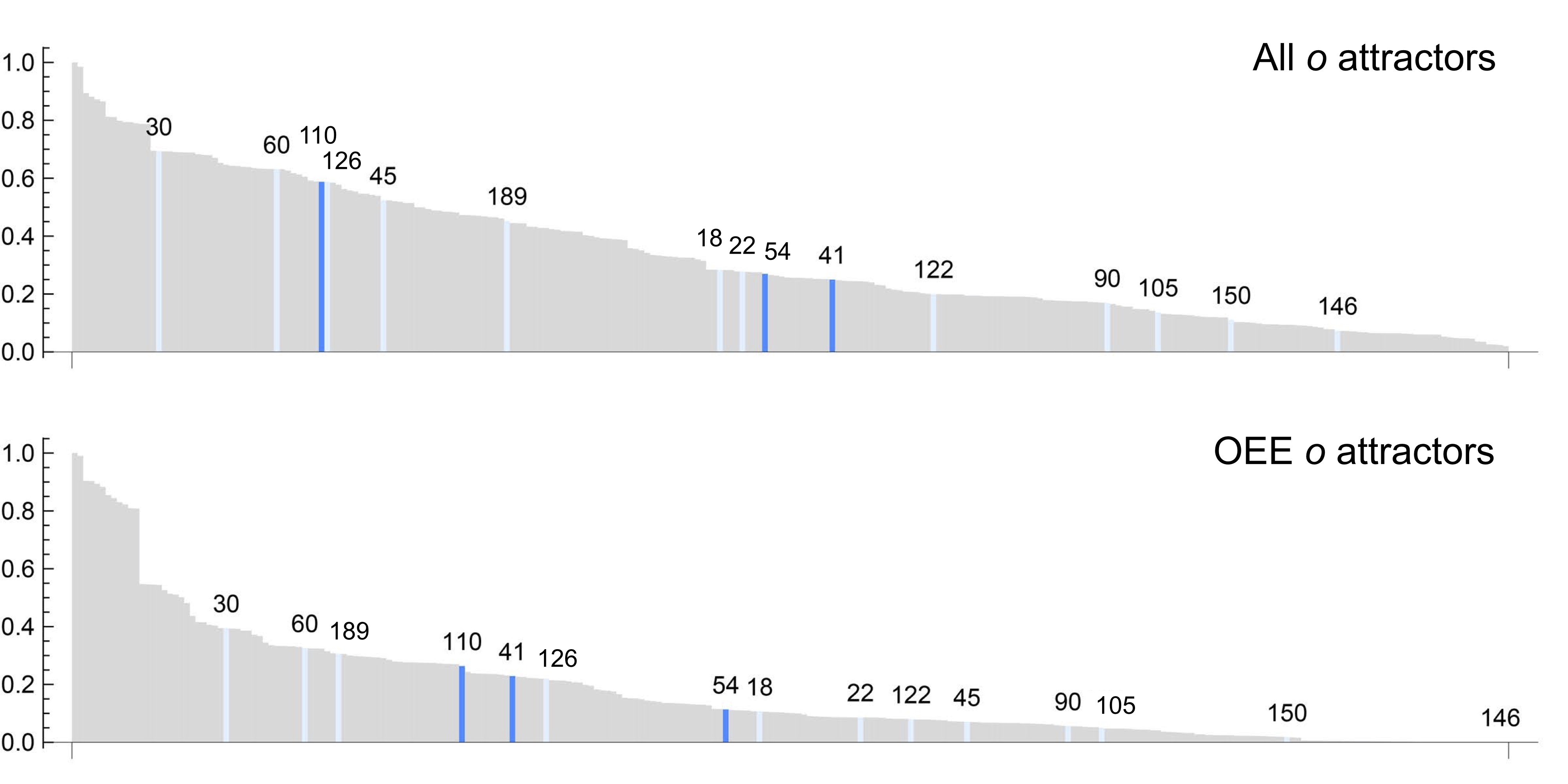}
\caption{Rank ordered frequency distributions of rules implemented in the attractor dynamics of $o$ for all sampled Case I CA (top) and OEE cases only (bottom). Highlighted are Wolfram Class III (light blue) and IV rules (dark blue).} \label{fig:metagenome}
\end{figure*}

\section*{Discussion}
We have provided formal definitions of {\it unbounded evolution} (UE) and {\it innovation} (INN) that can be evaluated in any finite dynamical system, provided it can be decomposed into two interacting subsystems $o$ and $e$. Systems satisfying both UE and INN we expect to minimally represent mechanisms capable OEE. Testing the criteria for UE and INN against three different CA models with time-dependent rules reveals what we believe to be quite general mechanisms applicable to a broad class of OEE systems.

\subsection*{Mechanisms for OEE}

Our analysis indicates that there are potentially many time-dependent mechanisms that can produce OEE in a subsystem $o$ embedded within a larger dynamical system, but that some may be more interesting than others. An externally driven time-dependence for the rules of $o$ (Case II), while producing the highest statistics of OEE cases sampled for small $o$, does not provide a scalable mechanism for producing OEE with increasing system size, unless the structure of $o$ itself is fundamentally altered (such that the rule space changes). Stochastically driven rule evolution displays rich transient dynamics, but ultimately subsystems converge on dynamics with low complexity (Case III). An alternative is to introduce stochasticity to the states, rather than the rules, which would avert this issue. This has the drawback that the mechanism for OEE is then not as clearly mappable to biological processes (or other mechanisms internal to the system), where the genotype (rules) evolve due to random mutations that then dictate the phenotype (states). 

We regard Case I as the most interesting mechanism explored herein for generating conditions favoring OEE: it is scalable and the dynamics generated are novel. We note that the state-dependent mechanism represents a departure from more traditional approaches to modeling dynamical systems, {\it e.g.} as occurs in the physical sciences, where the dynamical rule is usually assumed to be fixed. In particular, it represents an explicit form of top-down causation, often regarded as a key mechanism in emergence \cite{alglife, ellis2011top} that could also play an important role in driving major evolutionary transitions \cite{walker2012evolutionary}. The state-dependent mechanism is also consistent with an important hallmark of biology -- that biological systems appear to implement self-referential dynamics such that the ``laws'' in biology are a function of the states \cite{alglife, goldenfeld2011life, douglas1979godel}, a feature that also appears to be characteristic of the evolution of language \cite{languages, selfreference}.


\subsection*{Applicability to Other Dynamical Systems}
We have independently explored openness to an environment, stochasticity and state-dependent dynamics as we expect these to be general and apply to a wide-range of dynamical systems that might similarly display OEE by satisfying Definitions \ref{Def:UE} and \ref{Def:INN}. An important feature of these definitions is that UE and INN must be driven by {\it extrinsic} factors (an environment) \cite{taylor2004redrawing}, although the mechanisms driving the dynamics characteristic of OEE should be {\it intrinsic} to the subsystem of interest. OEE can therefore only be a property of a subsystem. We have not explored the case of feedback from $o$ to $e$ that might drive further open-ended dynamics, as characteristic of the biosphere, for example in niche construction \cite{laubichler2015extended}, but expect even richer dynamics to be observed in such cases. For large or infinite dynamical systems INN is an effective proxy for UE, and we expect highly innovative systems to be the most likely candidates for open-ended evolution. 

\section*{Conclusions}

Our results demonstrate that OEE, as formalized herein, is a general property of dynamical systems with time-dependent rules. This represents a radical departure from more traditional approaches to dynamics where the ``laws'' remain fixed. Our results suggest that uncovering the principles governing open-ended evolution and innovation in biological and technological systems may require removing the segregation of states and {\it fixed} dynamic laws characteristic of the physical sciences for the last $300$ years. In particular, state-dependent dynamics have been shown to out-perform other candidate mechanisms in terms of scalability, suggestive of paths forward for understanding OEE. Our analysis connects all four hallmarks of OEE and provides a mechanism for producing OEE that is consistent with the self-referential nature of living systems. By casting the formalism of OEE within the broader context of dynamical systems theory, the proof-of-principle approach presented opens up the possibility of finding unifying principles of OEE that encompass both biological and artificial systems.


\section*{Supplementary Materials}

\subsection*{Description of Implementations of Time-Dependent Cellular Automata Variants}\label{sec:MI}

We consider three new variants of cellular automata (CA) to identify mechanism(s) that can produce conditions necessary for open-ended evolution (OEE) in bounded regions, subject to the formal criteria for OEE laid out in Definitions 1 and 2 in the main text. We consider definitions of unbounded evolution (UE) and innovation (INN) that are applicable to {\it any} instance of a dynamical system $u$ that can be decomposed into two interacting subsystems $o$ and $e$. Each CA variant implements {\it time-dependent} rules for $o$, with different functional forms $f$ for this time-dependence. Here we describe in detail the implementation of each variant considered. 

\subsubsection*{Case I: Deterministic State-Dependent Rules in Subsystem $o$}

The first variant, Case I, implements {\it state-dependent} update rules. Case I CA are composed of two spatially separate, fixed-width, 1-dimensional CA: an ``organism'' $o$ and an environment $e$. Both $o$ and $e$ are implemented with periodic boundary conditions, and utilize the alphabet $\{0,1\}$.  The environment $e$ is an execution of an ECA, and is evolved according to a fixed rule drawn from the set of $256$ possible ECA rules, with periodic boundary conditions. 

The subsystem $o$ updates its rule according to a function $f$ such that $r_o(t+1) = f (s_o(t), r_o(t), s_e(t))$, where $s_o$ and $r_o$ are the state and rule of the organism and $s_e$ is the state of the environment. It is evolved with periodic boundary conditions. The expressed ECA rule of $o$ at time $t$, $r_o(t)$, is represented by the eight-bit binary representation of its rule table \cite{NKS}, {\it e.g.} an $o$ implementing Rule 30 at time $t$ will have $r_o(t) = [0,0,0,1,1,1,1,0]$ (see main text Fig. 3). We refer to individual bits within the rule by the index $i$ such that $r_o(t)[1] = 0$, $r_o(t)[2] = 0$, $r_o(t)[3] = 0$, $r_o(t)[4] = 1$ {\it etc.} for an $o$ implementing Rule 30 at time $t$. The binary representation of ECA rules are structured such that each successive bit $i$ iterated in this manner represents the output of application of the rule to the ordered set of triplet states $S^3=[111,110,101,100,011,010,001,000]$. 

The function $f$ for our example implementation of state-dependent CA is constructed such that at each time-step $t$ it compares the normalized frequency of each triplet $i$ in $S^3$ in the state of $o$ and $e$, $s_o(t)$ and $s_e(t)$, respectively, and flips the corresponding bit $i$ in $r_o(t)$ if $i$ is expressed in $s_o$ and the normalized frequency of the triplet in $s_o(t)$ meets or exceeds the normalized frequency in $s_e(t)$ (where the frequency is normalized relative to the number of possible triplets in the state). That is, at each time-step $t$, a bit $i$ in $r_o(t)$ will flip $0 \leftrightarrow 1$ if $n_i (s_o(t)) \geq n_i (s_e(t))$, where $n_i$ counts the relative frequency of triplet $i$. Formally, 

\begin{eqnarray} \label{eqn:f}
r_o(t+1) [i] = \begin{cases}
	      \overline{r_o(t) [i]} ~~~~ \text{if} ~~~ n_i(s_o(t)) \geq n_i(s_e(t)) \\
               r_o(t) [i] ~~~~ \text{if} ~~~ n_i(s_o(t)) < n_i(s_e(t)) \\
               \end{cases}
\end{eqnarray}
where the overbar represents logical negation.  

An example implementation of this update function is shown in Fig. 4 in the main text, where an ``organism'' $o$ with $w_o = 4$ is coupled to an environment $e$ with $w_e = 6$, and $r_o(t) = [0,0,0,1,1,1,1,0]$. In the example, only for $i=3$, corresponding to the triplet $\{1,0,1\}$, is  $n_3 (s_o(t)) \geq n_3 (s_e(t))$. Therefore, $r_o(t + 1) [3] =  \overline{r_o(t) [3]} = \bar{0} = 1$, as shown schematically in Fig. 5 in the main text. In this example, the interaction of $o$ and $e$ under $f$ changes $r_o$ from Rule 30 at time-step $t$ to Rule 62 at $t+1$.

\subsubsection*{Case II: Deterministic Time-Dependent Rules in Subsystem $o$}
The second variant, Case II, is similarly composed of two spatially separate, fixed-width, 1-dimensional CA: an ``organism'' $o$ and an environment $e$. As with Case I, both $o$ and $e$ are implemented with periodic boundary conditions, and utilize the alphabet $\{0,1\}$.  The environment $e$ is an execution of an ECA, and is evolved according to a fixed rule drawn from the set of $256$ possible ECA rules, just as in Case I. 

The key difference between Case I and Case II CA is that for Case II, the subsystem $o$ updates its rule according to a function $f$ such that $r_o(t+1) = f (s_e(t))$. That is, for Case II the update rule of $o$ depends {\it only} on the state of the external environment $s_e$ and is independent of the current state or rule of $o$ (that is, $o$ is {\it not} self-referential). Formally,
\begin{eqnarray}
r_o(t+1) [i] = s_e(t)[i]
\end{eqnarray}
Here $r_o(t)$ is determined uniquely by $s_e(t)$, such that the binary representation of each possible state of the environment uniquely maps to one ECA rule according to Wolfram's binary classification scheme \cite{NKS}. For this implementation the environment must be of width $w_e = 8$ to mediate a bijective map between $\{s_e\}$ and $\{r_o\}$. Case II CA emulate systems where the rules for dynamical evolution are modulated exclusively by the time evolution of an external system.

\subsubsection*{Case III: Stochastic Time-Dependent Rules in Subsystem $o$}

The final variant, Case III, is composed of a single, fixed-width, 1-dimensional CA: the ``organism'' $o$. Like Case II, the rule evolution of Case III is driven {\it externally} and does not depend on $s_o$. However, here the external environment $e$ is stochastic noise and not an ECA.  In Case III CA, the subsystem $o$ updates its rule according to a function $f$ such that $r_o(t+1) = f (r_o(t), \xi)$, where $\xi$ introduces random fluctuations in the implemented rule of $o$ by stochastically flipping bits in $r_o$. Formally,
\begin{eqnarray}
r_o(t+1) [i] = \begin{cases}
	      \overline{r_o(t) [i]} ~~~~ \text{if} ~~~ \xi < \mu \\
               r_o(t) [i] ~~~~ \text{if} ~~~ \xi \geq \mu \\
               \end{cases}
\end{eqnarray}
where $\mu$ is a fixed threshold for flipping between $[0,1)$, and $\xi$ is a random number drawn from the interval $[0,1)$.  This implements a diffusive-random walk through ECA rule space. Since the rule of $o$ at time $t+1$, $r_o(t+1)$, depends on the rule at time $t$, $r_o(t)$, the dynamics of Case III CA are path-dependent in a similar manner to Case I (both rely on flipping bits in $r_o(t)$, where Case I do so deterministically as a function of $s_o$ and $s_e$, and Case III do so stochastically).


\begin{table*}[h]
\centering
\caption{The size of the randomly sampled subspace for Case I CA for each $w_o$ and $w_e$ explored.}
\label{compstats1}
\begin{tabular}{llll|llll}
\textbf{CA Variant} & \textbf{$w_o$} & \textbf{\#u} & \textbf{\% Explored} & \textbf{CA Variant} & \textbf{$w_o$} & \textbf{\#u} & \textbf{\% Explored} \\ \hline
\begin{tabular}[c]{@{}l@{}}Case I:\\ $w_e = \frac{1}{2} w_o$\end{tabular} & 3 & 2.1\e{6} & 1.25 & \begin{tabular}[c]{@{}l@{}}Case I:\\ $w_e = 2 w_o$\end{tabular} & 3 & 3.36\e{7} & 6.92\e{-2} \\
 & 4 & 4.19\e{6} & 1.25 &  & 4 & 2.68\e{8} & 1.73\e{-2} \\
 & 5 & 1.68\e{7} & 0.62 &  & 5 & 2.15\e{9} & 4.69\e{-3} \\
 & 6 & 3.36\e{7} & 0.62 &  & 6 & 1.72\e{10} & 3.16\e{-4} \\
 & 7 & 1.34\e{8} & 0.31 &  & 7 & 1.37\e{11} & 7.75\e{-5} \\
\begin{tabular}[c]{@{}l@{}}Case I:\\ $w_e = w_o$\end{tabular} & 3 & 4.19\e{6} & 0.63 & \begin{tabular}[c]{@{}l@{}}Case I:\\ $w_e = \frac{5}{2} w_o$\end{tabular} & 3 & 6.71\e{7} & 3.91\e{-2} \\
 & 4 & 1.68\e{7} & 0.31 &  & 4 & 1.074\e{9} & 4.88\e{-2} \\
 & 5 & 6.71\e{7} & 0.16 &  & 5 & 8.59\e{8} & 1.22\e{-3} \\
 & 6 & 2.68\e{8} & 7.81\e{-2} &  & 6 & 1.37\e{11} & 1.53\e{-4} \\
 & 7 & 1.07\e{9} & 3.91\e{-2} &  & 7 & 1.1\e{12} & 3.81\e{-5} \\
\begin{tabular}[c]{@{}l@{}}Case I:\\ $w_e = \frac{3}{2} w_o$\end{tabular} & 3 & 8.34\e{6} & 0.28 &  &  &  &  \\
 & 4 & 6.71\e{7} & 6.92\e{-2} &  &  &  &  \\
 & 5 & 2.68\e{8} & 3.75\e{-2} &  &  &  &  \\
 & 6 & 2.15\e{9} & 2.52\e{-3} &  &  &  &  \\
 & 7 & 8.59\e{9} & 1.24\e{-3} &  &  &  & 
\end{tabular}
\end{table*}

\subsection*{Experimental Methods} 

The number of possible executions grows exponentially large with $w_o$, limiting the computational tractability of statistically rigorous sampling of the dynamics of each CA variant and of generating the set of counterfactual isolated ECA trajectories. We therefore explored small CA with $w_o = 3, 4,\ldots 7$ and sampled a representative subset of all possible trajectories for each $w_o$ (see Section \ref{sec:largeO} for examples of larger CA). We then generated statistics on the number of sampled trajectories satisfying Definitions 1 and 2 for unbounded evolution and innovation, respectively.

\subsubsection*{Case I Experiments} 

For Case I, $w_e$ must be specified in addition to $w_o$. We consider systems with $w_e = \frac{1}{2}w_o$, $w_o$, $\frac{3}{2}w_o$, $2w_o$ and $\frac{5}{2}w_o$ . For comparison to Case II and Case III CA, $w_e = w_o$ statistics are used. For each $w_o$ and $w_e$, the initial state of $o$, $s_o(0)$, the initial state of $e$, $s_e(0)$, the initial rule of $o$, $r_o(0)$ and the rule of $e$, $r_e$, are drawn at random. For $r_0(0)$ and $r_e$, we only consider the $88$ non-equivalent ECA rules, which  dramatically reduces the number of possible cases, but still covers the full spectrum of complexity in initial configurations. We then permit $r_o$ to evolve into any of the $256$ possible ECA rules. We also ensure that no two cases sampled are initialized with the same tuple $\{s_o(0), s_e(0), r_o(0), r_e\}$.

The space of all possible Case I CA executions is too large to explore the full space computationally. 
Since each $e$ and $o$ are each initiated with a state and a rule, the number of possible executions is:

\begin{equation}
N_U = N_{R}^2 \times N_{S_e} \times N_{S_o} = 88^2 \times 2^{8 w_e} \times 2^{8 w_o}
\end{equation}

\noindent where $N_R$ is the number of sampled initial rules for $o$ and $e$, $N_{S_e}$ is the number of sampled initial states for $e$, and $N_{S_o}$ is the number of sampled initial states for $o$.  For $w_o = w_e = 3$ and $w_o = w_e = 4$, exploring the full space of all possible initial conditions is computationally tractable, and verifies that the statistics reported herein for executions of $o$ that display UE and INN are characteristic of the full computational space for the smaller sample sizes implemented in this study. The number of randomly sampled cases for Case I CA included herein is given in Table \ref{compstats1}.

\subsubsection*{Case II Experiments} 

For Case II, $w_e = 8$ for all simulations, since this permits a bijective map from $\{s_e\}$ to the rule space of ECA and thus the set of rules $\{r_o\}$. As with Case I CA, executions are initialized with a randomized tuple $\{s_o(0), s_e(0), r_o(0), r_e\}$, ensuring that no two experiments are initialized with the same tuple.  We restrict attention only to $w_e = 8$ for Case II experiments in this study to directly compare to our Case I and Case III CA.  The number of randomly sampled cases for Case II CA is given in Table \ref{compstats2}.

\begin{table}[h!]
\centering
\caption{The size of the randomly sampled subspace for Case II CA for each $w_o$ explored.}
\label{compstats2}
\begin{tabular}{llll}
\textbf{CA Variant} & \textbf{$w_o$} & \textbf{\#u} & \textbf{\% Explored} \\ \hline
\begin{tabular}[c]{@{}l@{}}Case II:\\ $w_e = 8$\end{tabular} & 3 & 1.34\e{8} & 2.31\e{-2} \\
 & 4 & 2.68\e{8} & 2.1\e{-2} \\
 & 5 & 5.37\e{8} & 2.1\e{-2} \\
 & 6 & 1.07\e{9} & 2.1\e{-2} \\
 & 7 & 2.15\e{9} & 1.98\e{-2}
\end{tabular}
\end{table}

\subsubsection*{Case III Experiments} 

For Case III, a threshold $\mu$ for stochastic flipping of the bits in the rule table of $o$ must be set. Results for Case III are given for $\mu = 0.5$ in the main paper, such that each outcome bit in the rule table at every time step $r_o(t)$ has a 50\% probability of flipping. Results for other values are reported in Section \ref{sec:random}. Since we evolve only the subsystem $o$ for Case III CA, executions are initialized with a random tuple $\{s_o(0), r_o(0)\}$. We do not restrict sampled executions to unique tuples, since a different random seed is set for each execution. 
The number of randomly sampled cases for Case III CA is given in Table \ref{compstats3}. 

\begin{table}[h!]
\centering
\caption{The size of the randomly sampled subspace for Case III CA for each $w_o$ explored.} \label{compstats3}
\begin{tabular}{llll}
\textbf{CA Variant} & \textbf{$w_o$} & \textbf{\#u} & \textbf{\% Explored} \\ \hline
\begin{tabular}[c]{@{}l@{}}Case III:\\ Random\end{tabular} & 3 & 5.24\e{5} & 10 \\
 & 4 & 1.05\e{6} & 5 \\
 & 5 & 2.1\e{6} & 5 \\
 & 6 & 4.19\e{6} & 5 \\
 & 7 & 8.39\e{6} & 5
\end{tabular}
\end{table}


\subsection*{Calculating Recurrence Time, Compressibility and Lyapunov Exponent}\label{sec:calc}

Recurrence times for the state- and rule-trajectory of $o$ were calculated to identify cases exhibiting UE and thus OEE. The complexity of the state trajectory $\{s_o(0), s_o(1), \ldots s_o(t_r)\}$ was measured by means of its {\it compressibility} ($C$), and calculation of the {\it Lyapunov exponent} ($k$). 

\subsubsection*{Recurrence Time}\label{sec:rectimes}

\begin{figure}[h!]
    \centering
    \includegraphics[width=0.15\textwidth]{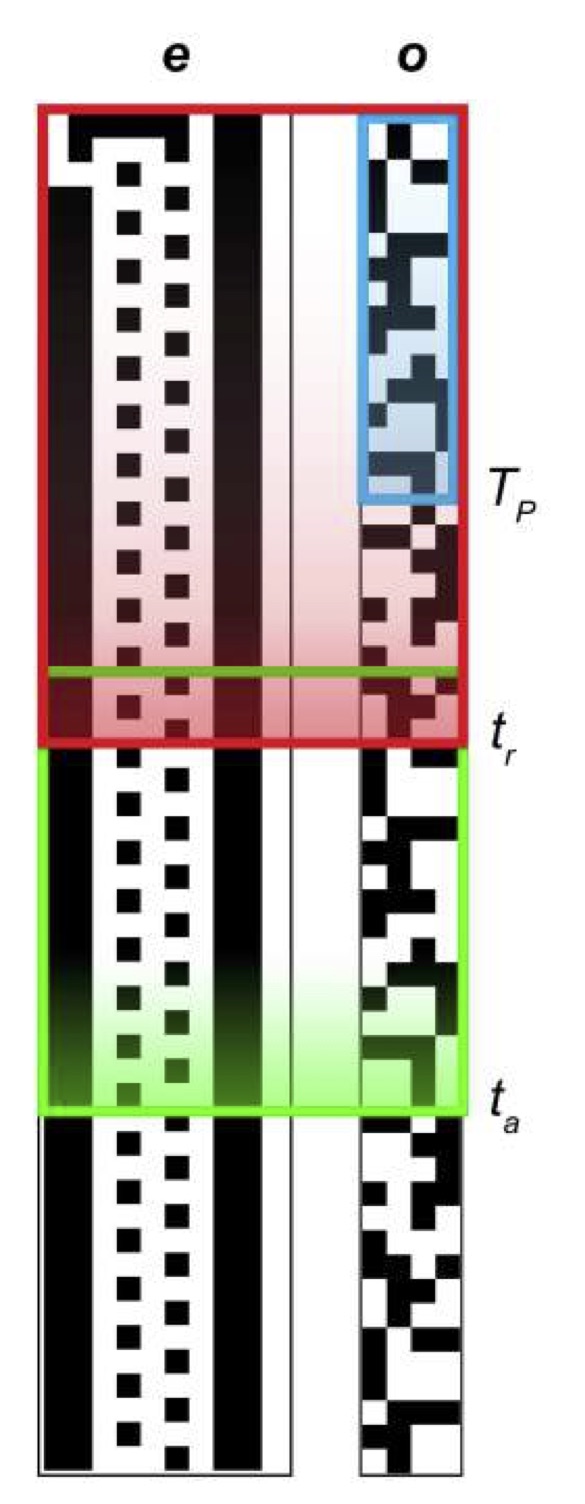}
    \caption{Relevant timescales for describing the dynamics of $o$ embedded in $u$. Shown are the Poincar\'{e} recurrence time $t_P$ (blue) for an isolated ECA of the same width $w_o$ as $o$, the state-trajectory recurrence time $t_r$ of $o$ (red), and attractor size of the full system $u$, $t_a$ (green).}
    \label{fig:example_ecosystem}
\end{figure}

For Cases I and II,we measured the recurrence times $t_r'$ and $t_r$ for $o$, for both the rule evolution $\{ r_o(t_1),r_o(t_2),r_o(t_3) \ldots r_o(t_r') \}$ and the state evolution $\{ s_o(t_1),s_o(t_2),s_o(t_3) \ldots s_o(t_r) \}$, respectively. Recurrence times were calculated by determining the time $t_r$ or $t_r'$ when the {\it sequence} of states or rules of $o$, respectively, repeated. In general, $t_r$ and $t_r'$ for $o$ are not the same as for the full system $u$ (or as each other, such that often $t_r \neq t_r'$, see Fig. 1 in the main text). We therefore first determined when $u$ repeated the tuple $\{s_o, s_e, r_o\}$ such that $\{s_o(t'), s_e(t'), r_o(t')\}$ = $\{s_o(t), s_e(t), r_o(t)\}$, where $t < t'$. We then determined the $t_r$ such that $\{s_o(t_r), s_o(t_r + 1), \ldots s_o(t')\} = \{s_o(t_i), s_o(t_i + 1), \ldots s_o(t_r)\}$ for $t_i < t_r$ (and likewise for $t_r'$ with the replacement $r_o$ for $s_o$). The time step $t_i$ is identified as initiation of the attractor dynamics for $o$. In many cases, we find attractors that are unbounded and innovative by Definitions 1 and 2, in addition to full trajectories up to recurrence. An example illustrating the expected Poincar\'e time for $o$, $t_P$, its recurrence time for the state trajectory $t_r$ and the attractor size for the full system $u$, $t_a$ (up to the recurrence time $t'$ for the full system) is shown in Fig. \ref{fig:example_ecosystem}. 

Since Case III CA are stochastically evolved, their dynamics do not repeat with a unique recurrence time $t_r$ for $o$. However, all executions sampled eventually terminated in an oscillation between the two homogeneous states (all-`0's or all-`1's). These states are attractors for every fixed rule ECA evolved under periodic boundary conditions, so once a Case III CA evolves to either homogeneous state, no heterogeneity will ever be produced (the dynamics behave somewhat like dissipation of the heterogeneity in the initial state). We therefore consider it more meaningful to calculate the number of time steps before convergence to this oscillatory attractor in place of the recurrence time $t_r$, which we denote by $t_r$ for consistency of notation with other cases explored. We therefore capture the timescale of relevance for all interesting (and potentially complex) dynamics, which occur in the transient before converging to this attractor. 

\subsubsection*{Compressibility} The Kolmogorov-Chaitin complexity of string $s$ is defined as the size of the shortest computer program $p$ running on a universal Turing machine $U$ that produces the string $s$ (here $s$ is the sequence of states of $o$):
\begin{equation}
K_U(s) = min\{|p|, U(p) = s\}~.
\end{equation}
Although it cannot be computed exactly, it is lower semi-computable and can be approximated by using a general lossless compression algorithm $L$ \cite{compression}.
This upper-bound approximation of the Kolmogorov-Chaitin complexity is normalized according to a normalized compression measure $C$:
\begin{equation} \label{eqn: C}
C(s) = \frac{L(s)}{max(C_i(s),length(s))}~.
\end{equation}
Throughout this paper $C_i$ is output of the Compress algorithm based on the LZW algorithm~\cite{compression}. It can be replaced by the output of any other compression algorithm. The measure is therefore a family of possible indexes approximating $K$. We use $C$ as measure over the state-trajectory of the organism $o$ for each execution $u$, as an approximation of the characteristic complexity of $o$ in the limit of large times $t \rightarrow \infty$.

{\it Large values of $C$ indicate low Kolmogorov-Chaitin complexity, meaning the output can be produced by a simple (short) program $p$}. The normalization constant $max(C(s),length(s))$ was calculated by measuring the number of bits resulting from a generalized compression algorithm for the Poincar\'e recurrence time of the entire system $u$, not an isolated organism. This allows normalizing the observed $C$ to its maximum possible value for an organism coupled to an environment. This closely approximates an upper limit in $C$ for the longest possible non-repeating trajectory for any given $o$.

In order to ensure the normalization constant for an organism of width $w_o$ is a close approximation to the maximal value, $C_i(s)$ was calculated for  $10^7$ randomly generated ECA of width $w_o$, evolved with a fixed rule for $2^{2w}$ time steps, where $w$ is the width of $u$, such that $w = w_o + w_e$. The maximum of this set was used as the normalization constant $max(C_i(s),length(s))$. Thus, all $C$ values are normalized relative to the maximal complexity of a CA evolved according to a fixed dynamical rule. 

\subsubsection*{Lyapunov Exponent} The Lyapunov exponent $k$ captures the speed at which a perturbation moves through a system \cite{lyp}, thereby quantifying sensitivity to initial conditions. In CA, $k$ can be, in general, measured by perturbing a single bit in the initial condition, and counting how many bits differ compared to the unperturbed time evolution in each time step:
\begin{equation}
y(t) = H_{io}[s_i(t), s_o(t)]
\end{equation}
where $H_{io}$ is the Hamming distance between the state of the perturbed system $i$ and the original organism $o$, which is evaluated at each time step $t$. The resulting time series of $y(t)$ values can be approximated as an exponential function, $y(t) = e^{kt}$, where $k$ is estimated numerically. High values of $k$ indicate sensitivity to perturbations, which is typically associated with complex dynamical systems, such as those that occur in deterministic chaos.


\subsection*{Statistics of Sampled Trajectories Displaying Innovation (INN)}

Tables \ref{innotable1} and \ref{innotable2} show the resulting statistics for sampled $o$ that were found to be innovative (INN) according to Definition 2. Table \ref{innotable1} includes all three CA variants as well as ECA counterfactual trajectories used as a control. Table \ref{innotable2} shows results for state-dependent Case I CA as a function of varying environment size $w_e$. Innovative $o$ were identified as having a state-trajectory that cannot be reproduced by any closed, fixed rule ECA of equivalent width $w = w_o$.

\begin{table}[h!]
\centering
\caption{Percentage of sampled cases displaying INN for each CA variant.}
\label{innotable1}
\begin{tabular}{l|llll}
\textbf{$w_o$} & \textbf{ECA} & \textbf{Case I ($w_o = w_e$)} & \textbf{Case II} & \textbf{Case III} \\ \hline
3 & 0 & 54.62 & 99.98 & 99.82 \\
4 & 0 & 74.66 & 99.97 & 99.87 \\
5 & 0 & 92.56 & 99.97 & 99.92 \\
6 & 0 & 88.14 & 99.97 & 99.94 \\
7 & 0 & 97.14 & 99.97 & 99.97
\end{tabular}
\end{table}

\begin{table}[h!]
\centering
\caption{Percentage of sampled cases displaying INN for Case I, with varying environment size $w_e$.}
\label{innotable2}
\begin{tabular}{l|lllll}
\textbf{$w_o$} & \textbf{$w_e = \frac{1}{2}w_o$} & \textbf{$w_e = w_o$} & \textbf{$w_e = \frac{3}{2}w_o$} & \textbf{$w_e = 2w_o$} & \textbf{$w_e = \frac{5}{2}w_o$} \\ \hline
3 & 30.72 & 54.62 & 70.10 & 86.04 & 93.29 \\
4 & 33.32 & 74.66 & 86.57 & 95.52 & 97.47 \\
5 & 32.42 & 92.56 & 96.22 & 98.32 & 98.72 \\
6 & 35.64 & 88.14 & 97.03 & 98.91 & 99.29 \\
7 & 52.92 & 97.14 & 97.43 & 99.51 & 99.63
\end{tabular}
\end{table}

\subsection*{Recurrence time frequency distributions}\label{sec:random}

\begin{figure}[h!]
\centering
\includegraphics[width=1\textwidth]{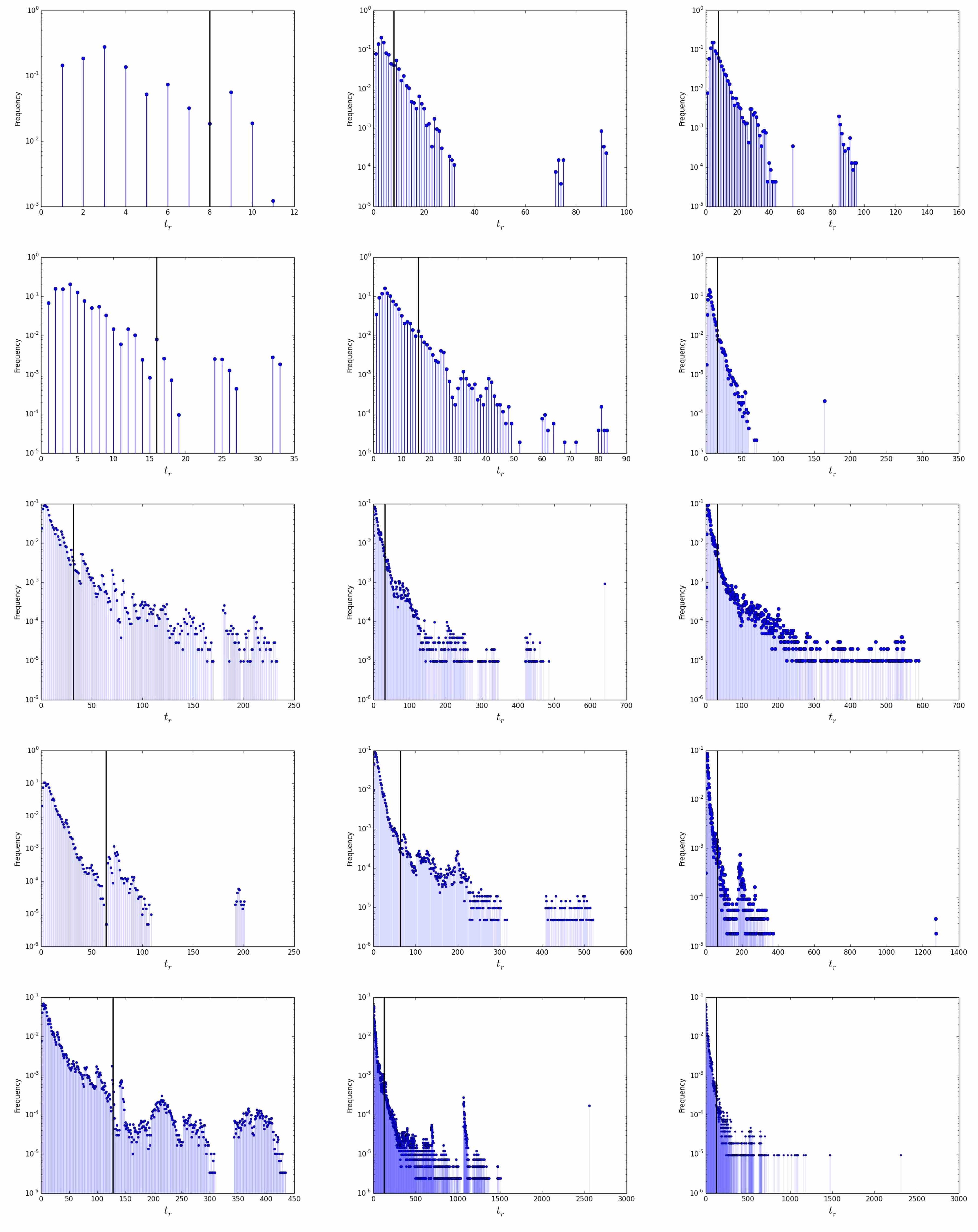}
\caption{Frequency distributions of recurrence times $t_r$ for Case I CA with $w_e = w_o$ (leftmost column), $w_e = \frac{3}{2} w_o$ (left middle), $w_e = 2 w_o$ (right middle) and $w_e = \frac{5}{2} w_o$ (rightmost column). For rows from top to bottom, $w_o = 3, 4, 5, 6$ and $7$ respectively. The Poincar\'e recurrence time $t_P$ of an isolated ECA of width $w_o$ is highlighted by the black vertical line in each panel.}
\label{fig:freq_CaseI}
\end{figure}

\begin{figure}[h!]
    \centering
    \includegraphics[width=\textwidth]{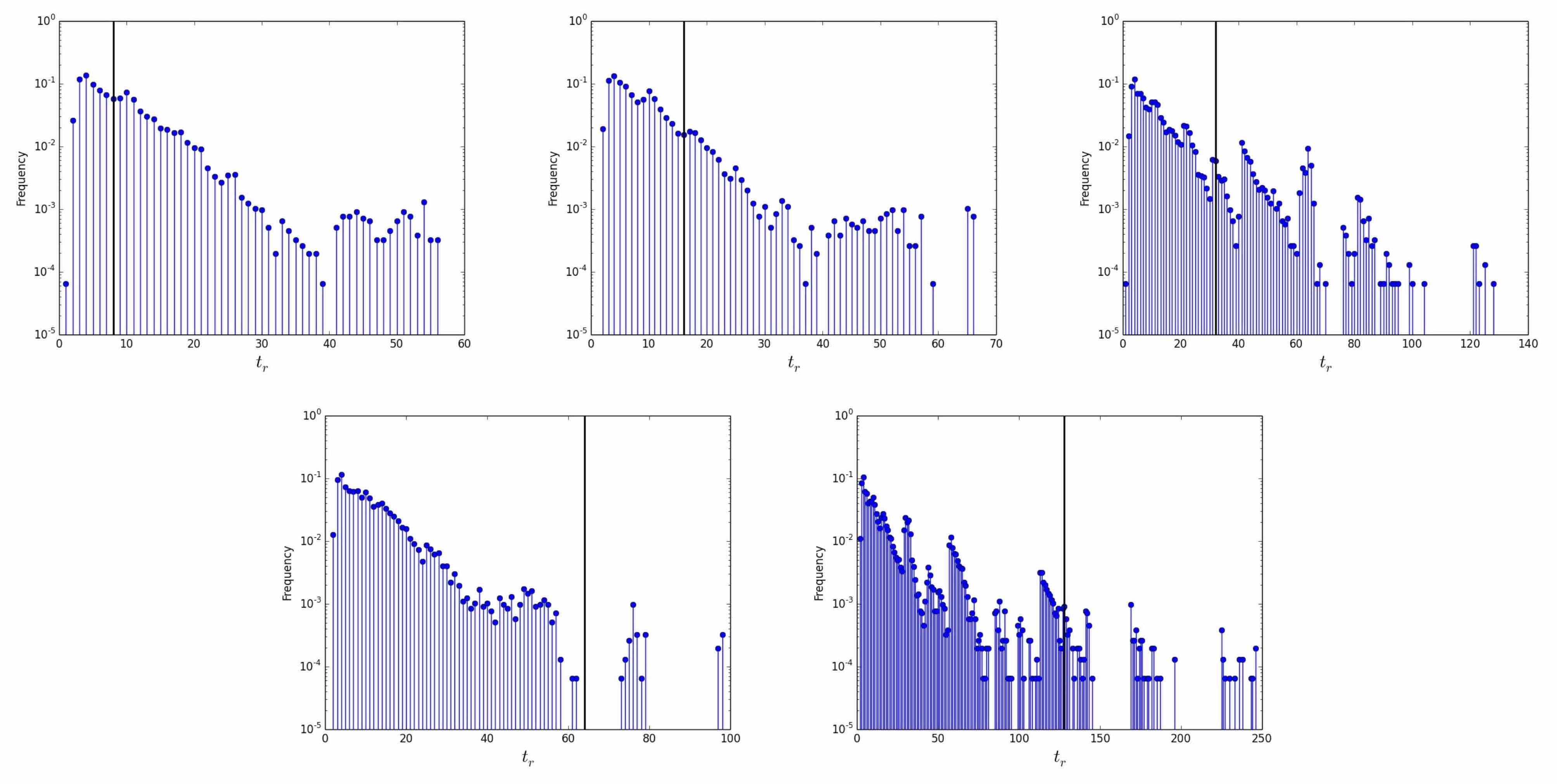}
    \caption{Frequency distributions of recurrence times $t_r$ for Case II CA. From top to bottom, $w_o = 3, 4, 5, 6$ and $7$ respectively. The Poincar\'e recurrence time $t_P$ of an isolated ECA of width $w_o$ is highlighted by the black vertical line in each panel.}
    \label{fig:freq_CaseII}
\end{figure}

\begin{figure}[h!]
\centering
\includegraphics[width=\textwidth]{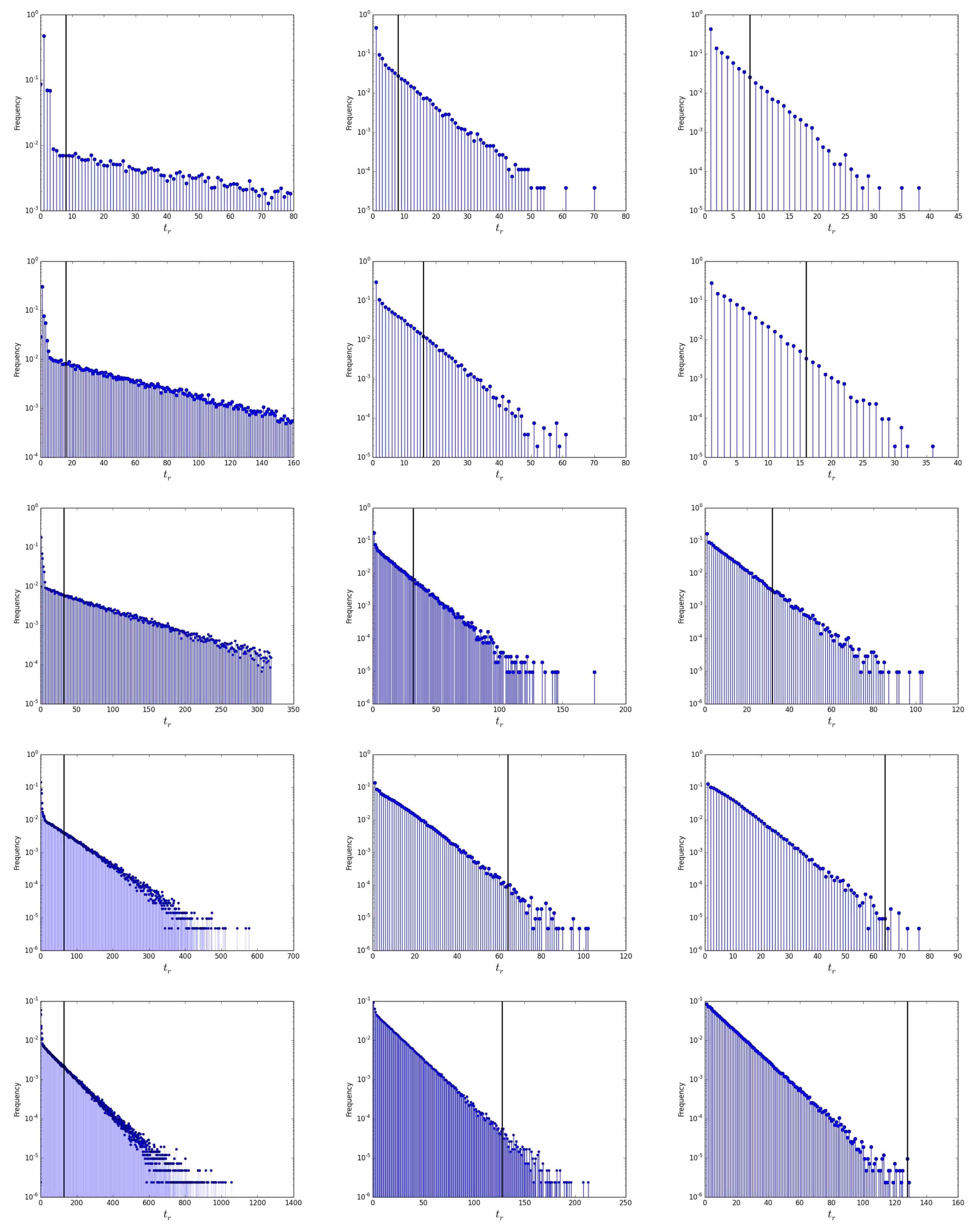}
\caption{Frequency distributions of $t_r$ for Case III CA with $\mu = 0.01$ (left), $\mu = 0.1$ (middle) and $\mu = 0.5$ (right). From top to bottom, $w_o = 3, 4, 5, 6$ and $7$ respectively. The Poincar\'e recurrence time $t_P$ of an isolated ECA of width $w_o$ is highlighted by the black vertical line in each panel.}
\label{fig:freq_CaseIII}
\end{figure}

The frequency distribution of $t_r$ observed for sampled state trajectories of $o$ in Case I, Case II and Case III CA are shown in Figures \ref{fig:freq_CaseI},\ref{fig:freq_CaseII} and \ref{fig:freq_CaseIII}, respectively. Comparing the three cases reveals that for equivalently sized ensembles of sampled trajectories for Case I, Case II, and Case III CA, the Case I CA generate OEE cases with higher statistical certainty than either the Case II or Case III CA for most parameters explored. This is especially true for cases where $w_e > w_o$ in Case I simulations. From Figure \ref{fig:freq_CaseI} it is evident that larger environments yield more UE cases with $t_r > t_P$ and in general result in longer observed recurrence times.

Case II CA yield fewer OEE cases as $w_o$ increases, as evident in Figure \ref{fig:freq_CaseII}. As discussed in the main text, Case II is not scalable as it would require changing the structure of the rules of the organism $o$. 

Case III CA generate fewer OEE cases than Case I as the width of $o$ increases, with no cases observed in our statistical sample for $w_o > 7$ for $\mu = 0.5$ (Figure \ref{fig:freq_CaseI} bottom panel, leftmost column). The frequency distribution of recurrence times for Case III CA with $\mu = 0.01$, $\mu = 0.1$ and $\mu = 0.5$ are shown in Figure \ref{fig:freq_CaseI}. For smaller values of $\mu$ OEE cases are observed for larger $w_o$. In the context of biological evolution,  the mechanism for increasing the number of OEE cases under Case III would therefore be for systems to evolve toward {\it slower} mutation rates over time. However, because the distributions are exponentially distributed, the number of OEE cases is always exponentially suppressed, representing only a small tail of the distribution. A fixed width $w_o$ execution could always be found in a {\it large enough} statistical sample that such that the observed $t_r$ would be greater than the maximum recurrence time observed in a Case I CA with an equivalent organism width $w_o$. However, in general due to the exponential suppression of cases with larger $t_r$ for Case III variants (Figure~\ref{fig:freq_CaseIII}), the ensemble size of sampled cases will necessarily be much larger for Case III CA than for Case I CA. That is, for a sufficiently large ensemble size one could chose a value of $\mu$ and generate a trajectory in a width $w_o$ organism with a given recurrence time $t_r$, but would always be able to find an example trajectory of the same $t_r$ for a {\it smaller} sized ensemble of Case I CA for some environment width $w_e$. Due to the exponential suppression, OEE cases are much rarer for Case III than Case I CA. Additionally, once Case III reach the terminal attractor their dynamics are not complexity, whereas Case I CA will repeat an attractor state that is in general complex and is often times open-ended (such that the attractor itself satisfies Definitions 1 and 2). We therefore regard Case III to not be scalable. 

\newpage

\newpage

\newpage

\newpage
\subsection*{ECA Rule Complexity of Case I CA}

To determine if the complexity observed in Case I CA is {\it intrinsic} to the state-dependent mechanism, or is an artifact of a selection-effect favoring complex rules, we determined the frequency of rules implemented in Case I CA utilizing the Wolfram classification scheme for Elementary Cellular Automata \cite{NKS}. There are four Wolfram Classes: Class I and II are regarded as the least complex, often generating simple repeating patterns. Class III rules are more complex displaying random patterns, and Class IV are regarded as the most complex, displaying rich dynamical structure (for example, ECA Rule 110, which is known to be Turing Universal~\cite{110} is a Class IV ECA). We analyze the complexity of ECA rules implemented in the rule trajectories of Case I CA by considering the frequency of implementation of rules from each class to determine if the complexity of the observed dynamics is an artifact of the ECA rules or {\it intrinsic} to $f$.

\begin{figure*}[ht]
    \centering
    \includegraphics[width=1\textwidth]{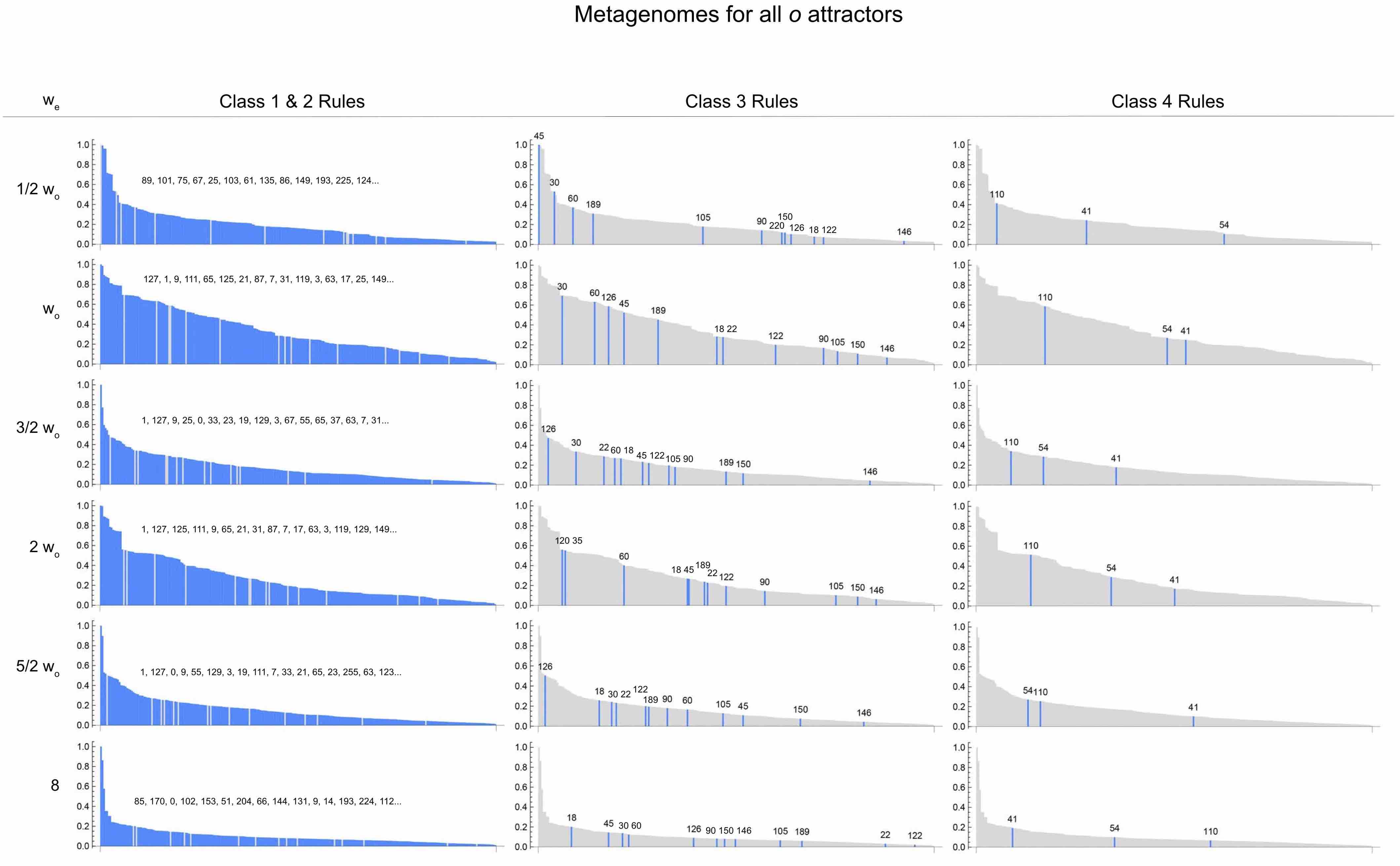}
    \caption{Rank ordered frequency distributions of rules (``metagenomes'') implemented by $o$ in its attractor. From top to bottom $w_e = \frac{1}{2}w_o$, $w_o$, $\frac{3}{2}w_o$, $2w_o$ and $\frac{5}{2}w_o$, respectively. Highlighted in blue are the frequencies of Class I and II rules (left), Class III rules (middle) and Class IV rules (right).}
    \label{meta2} 
\end{figure*}

\begin{figure*}[h]
    \centering
    \includegraphics[width=1\textwidth]{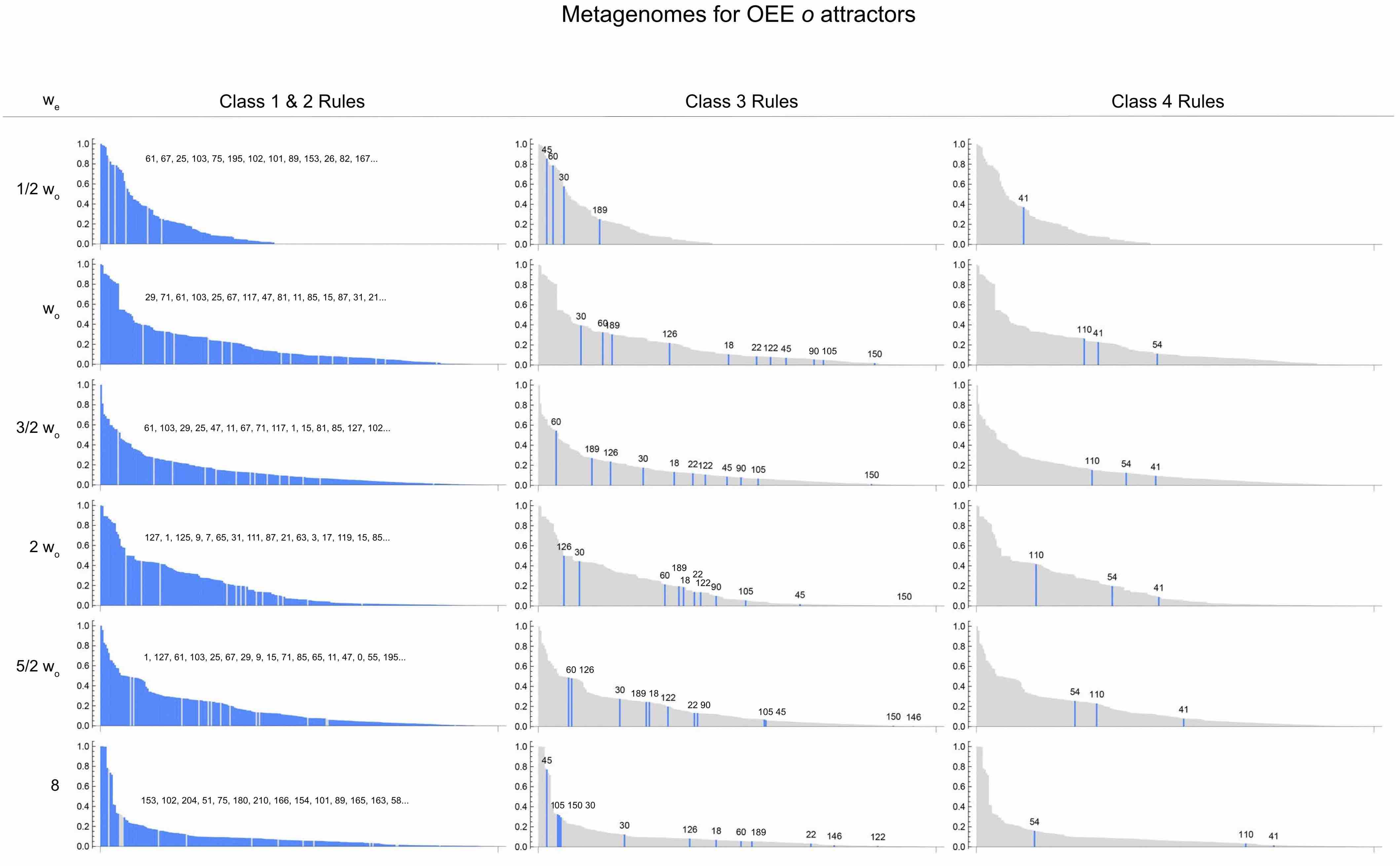}
    \caption{Rank ordered frequency distributions of rules (``metagenomes'') implemented by $o$ in its attractor for OEE cases only. From top to bottom $w_e = \frac{1}{2}w_o$, $w_o$, $\frac{3}{2}w_o$, $2w_o$ and $\frac{5}{2}w_o$, respectively. Highlighted in blue are the frequencies of Class I and II rules (left), Class III rules (middle) and Class IV rules (right).}
    \label{meta1}
\end{figure*}

The resulting rank ordered frequency distribution of rules is shown in Figure~\ref{meta2} for all sampled Case I CA of a given organism width $w_o$, and separately for the OEE cases in Figure~\ref{meta1}. Since this data includes statistics for the entire sample of $o$ of a given width $w_o$ included in our study, we call these distributions ``metagenomes'' to indicate that they represent bulk statistics over many instances of ``organisms'' $o$. The resulting distributions indicate that Case I CA primarily implement Class I and II rules, indicative that the complexity observed is {\it intrinsic} to the state-dependent mechanism and not an artifact of selective use of `complex' Class III and IV ECA rules. This is true for statistics sampled over all Case I CA (Figure~\ref{meta2}), as well as isolating only OEE cases (Figure~\ref{meta1}).

\newpage

\newpage
\subsection*{Distributions of Attractor Sizes}\label{sec:attractors}

\begin{figure}[h]
\centering
\includegraphics[width=0.4\linewidth]{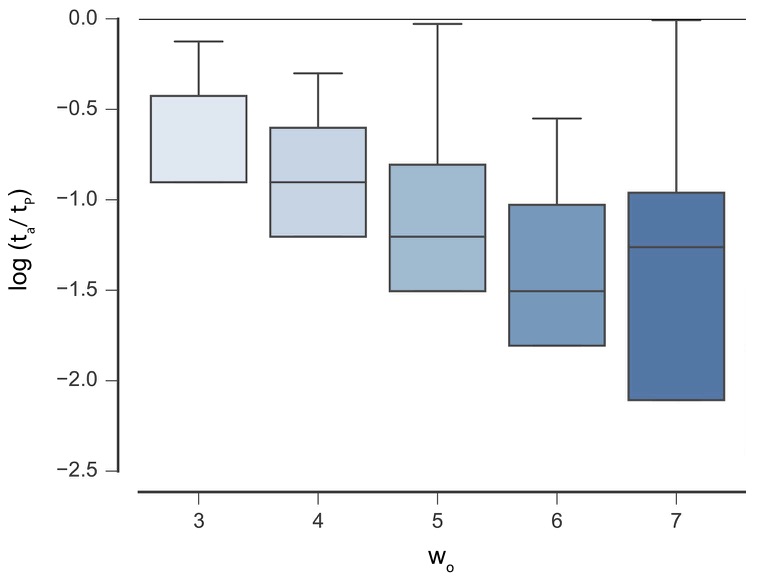}
\caption{Distribution of attractor sizes $t_a$ for the state trajectory of for all $88$ non-equivalent ECA rules, evolved from all possible initial conditions of width $w_o$. Attractor sizes are normalized to the Poincar\'e time $t_P = 2^{w_o}$ for an isolated ECA, where the black horizontal line indicates where $t_r/t_P = 1$ (shown on a log scale).} \label{fig:att_ECA}
\end{figure}

\begin{figure}[h]
\centering
\includegraphics[width=0.4\linewidth]{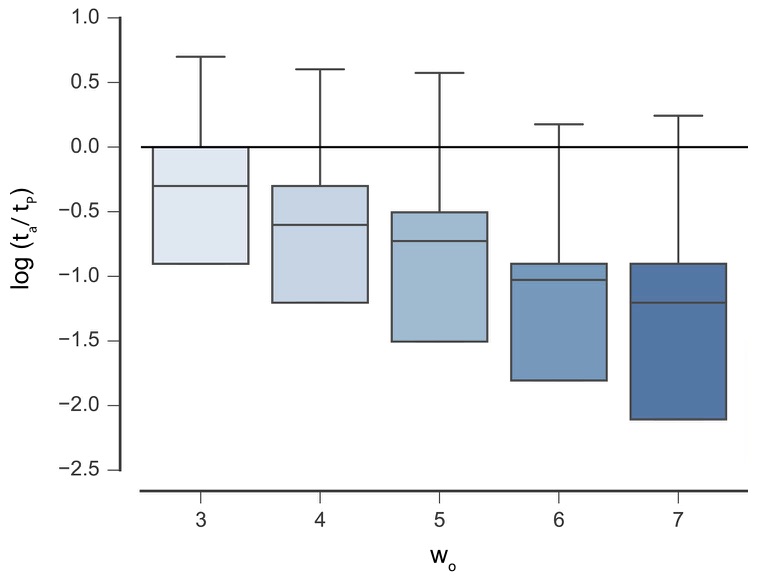}
\caption{Distribution of attractor sizes $t_a$ for the state trajectory of $o$, for Case II CA. Attractor sizes are normalized to the Poincar\'e time $t_P = 2^{w_o}$ for an isolated ECA. The black horizontal line indicates where $t_a/t_P = 1$ (shown on a log scale). Sample trajectories displaying {\it unbounded evolution} (UE) occur for $t_a/t_P > 1$.} \label{fig:att_CaseII}
\end{figure}

Figures~\ref{fig:att_ECA} - \ref{fig:att_CaseII} show box-whisker plots of the attractor sizes for the subsystem $o$ for each ECA and for Case I and Case II CA (Case III CA terminate in a random, oscillatory attractor and the statistics are therefore not included here, see Section \ref{sec:rectimes} for discussion). In each figure, the black horizontal line indicates where $t_r/t_P = 1$, where $t_a$ is the attractor size and $t_P$ is the expected Poincar\'e time of an equivalent isolated system (an ECA).  Sampled attractors exhibiting unbounded evolution (UE) have $t_a/t_P > 1$ and therefore fall above the black solid line - these are examples of OEE attractors.  

\newpage
\begin{figure}[ht]
\centering
\includegraphics[width=\linewidth]{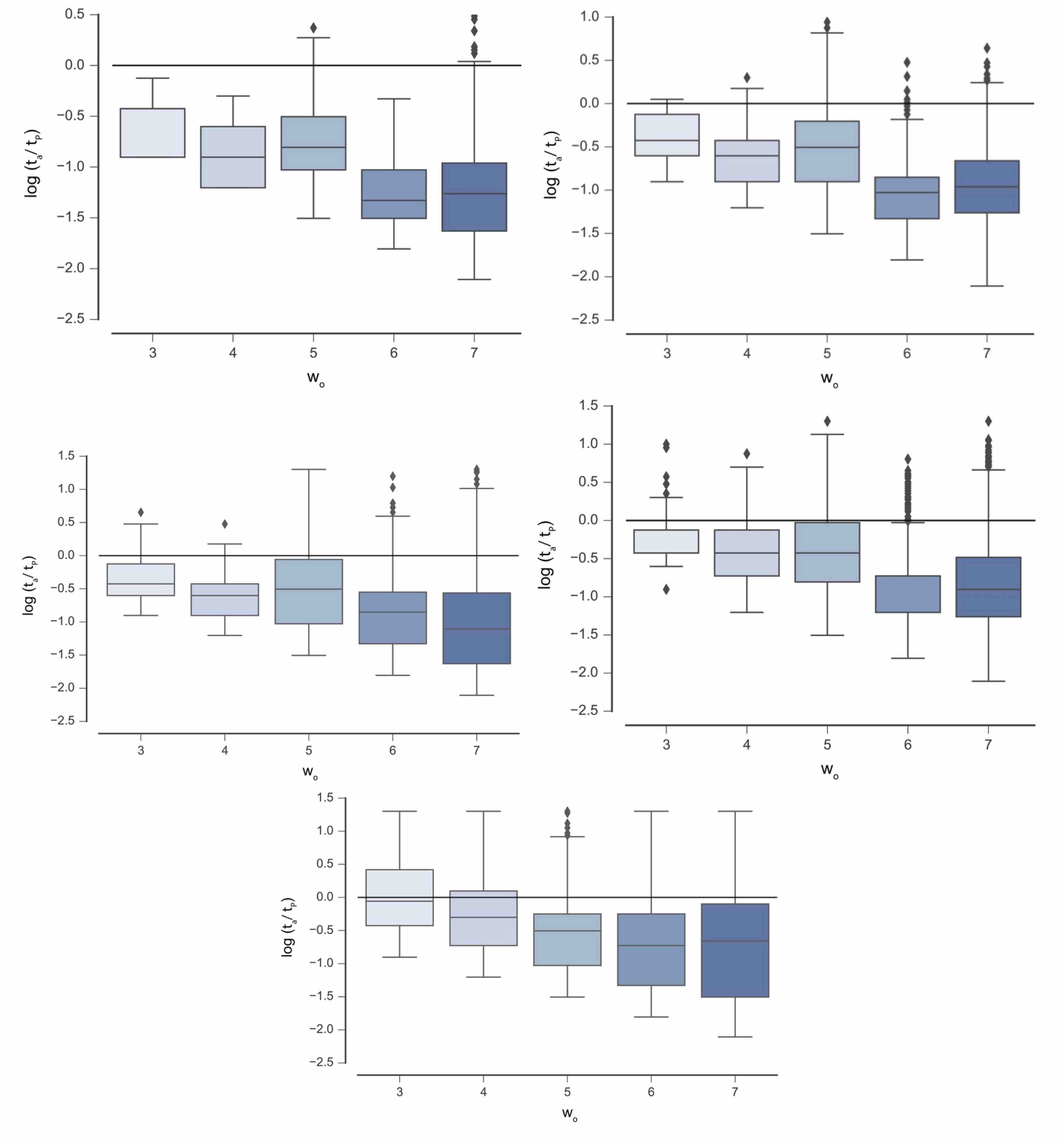}
\caption{Distribution of attractor sizes $t_a$ for the state trajectory of $o$ for Case I CA. Shown from top to bottom are distributions for $w_e = \frac{1}{2}w_o$, $w_o$, $\frac{3}{2}w_o$, $2w_o$ and $\frac{5}{2}w_o$, respectively. Attractor sizes are normalized to the Poincar\'e time $t_P = 2^{w_o}$ for an isolated ECA. 
The black horizontal line indicates where $t_a/t_P = 1$ (shown on a log scale). Sampled trajectories displaying UE occur for $t_r/t_P > 1$.}
 \label{fig:att_CaseI}
\end{figure}

\newpage
\subsection*{Compressibility and Lyapunov Exponent Values for Case I and II CA}


\subsubsection*{Compressibility and Lyapunov exponent for OEE trajectories sampled from Case I CA}

\begin{figure*}[h]
    \centering
    \includegraphics[width=\textwidth]{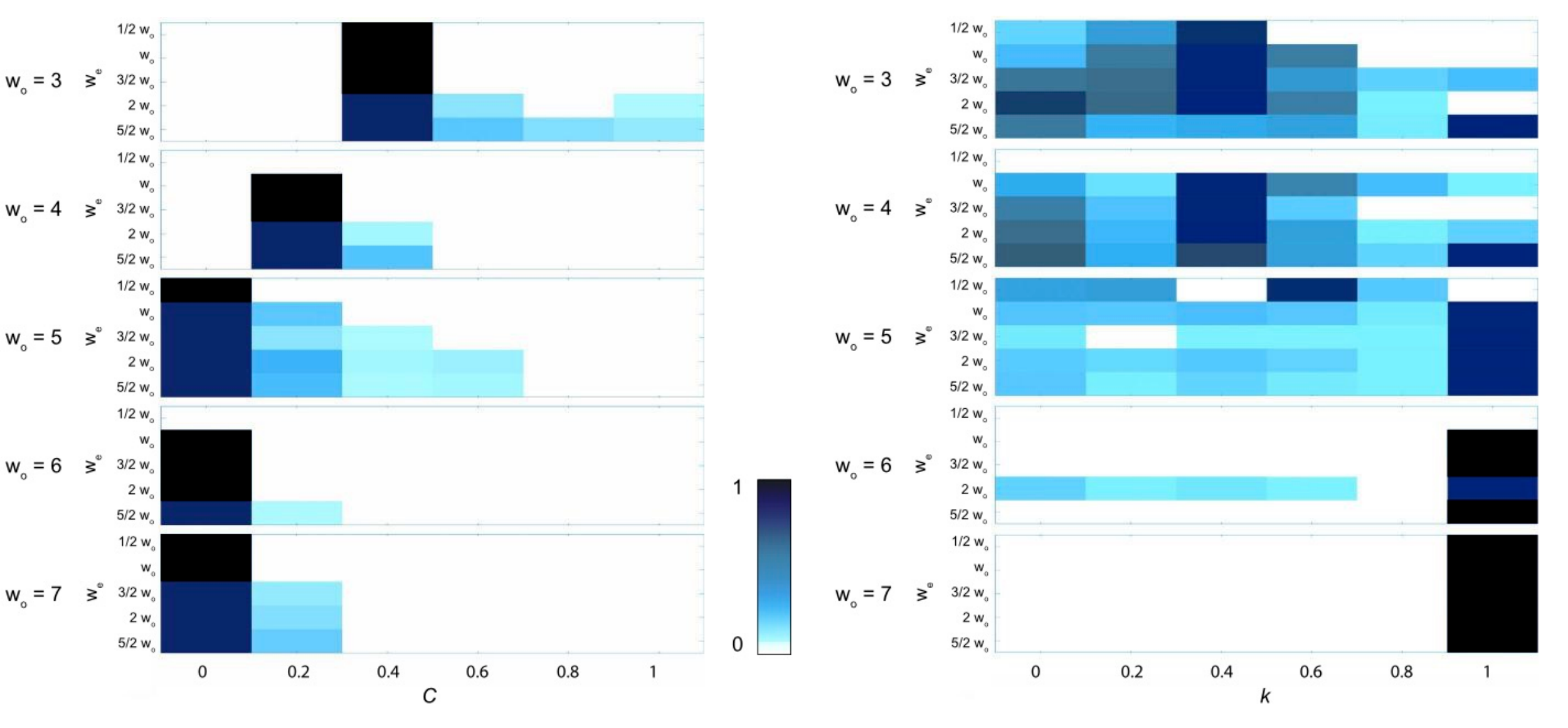}
    \caption{Heat maps of compression $C$ (left) and Lyapunov exponent values $k$ (right) for sampled OEE trajectories for the states of $o$ for Case I CA. From top to bottom $w_o  = 3, 4, 5, 6$ and $7$, with distributions shown for $w_e = \frac{1}{2}w_o$, $w_o$, $\frac{3}{2}w_o$, $2w_o$ and $\frac{5}{2}w_o$ (from top to bottom, respectively) for each $w_o$. Distributions are normalized to the total size of sampled trajectories for each $w_o$ and $w_e$ (see statistics in Table \ref{compstats1}).}
    \label{fig:CK_CaseIOEE}
\end{figure*}

Calculated values for compressibility ($C$) and Lyapunov exponent ($k$), as defined in Section~\ref{sec:calc}, are shown  in Figure \ref{fig:CK_CaseIOEE} for the state trajectory of $o$ for all sampled OEE executions for Case I CA. Comparison of the left panel of Figure \ref{fig:CK_CaseIOEE} with the left panel of Figure 2 in the main text reveals that the observed $C$ for all OEE cases tends to be lower than that calculated over all sampled $o$ for Case I: that is, the OEE cases exhibit lower $C$, consistent with intuition that systems with longer recurrence times should be 'more complex'. As $w_o$ increases, more OEE cases tend to have lower $C$ values, such that larger ``organisms'' are more complex. 

Likewise, comparing the right panel of Figure \ref{fig:CK_CaseIOEE} with the right panel of Figure 2 in the main text indicates that OEE cases also tend to have much higher $k$ values than that calculated over all sampled $o$ for Case I, indicative of greater sensitivity to perturbations in OEE systems. Large values of $w_o$ lead to larger $k$, on average, such that larger OEE ``organisms'' are more sensitive to perturbations and therefore display richer, more complex dynamics. 

\newpage
\subsubsection*{Comparison of compressibility and Lyapunov exponent for Case I and Case II CA}

\begin{figure*}[h]
    \centering
    \includegraphics[width=\textwidth]{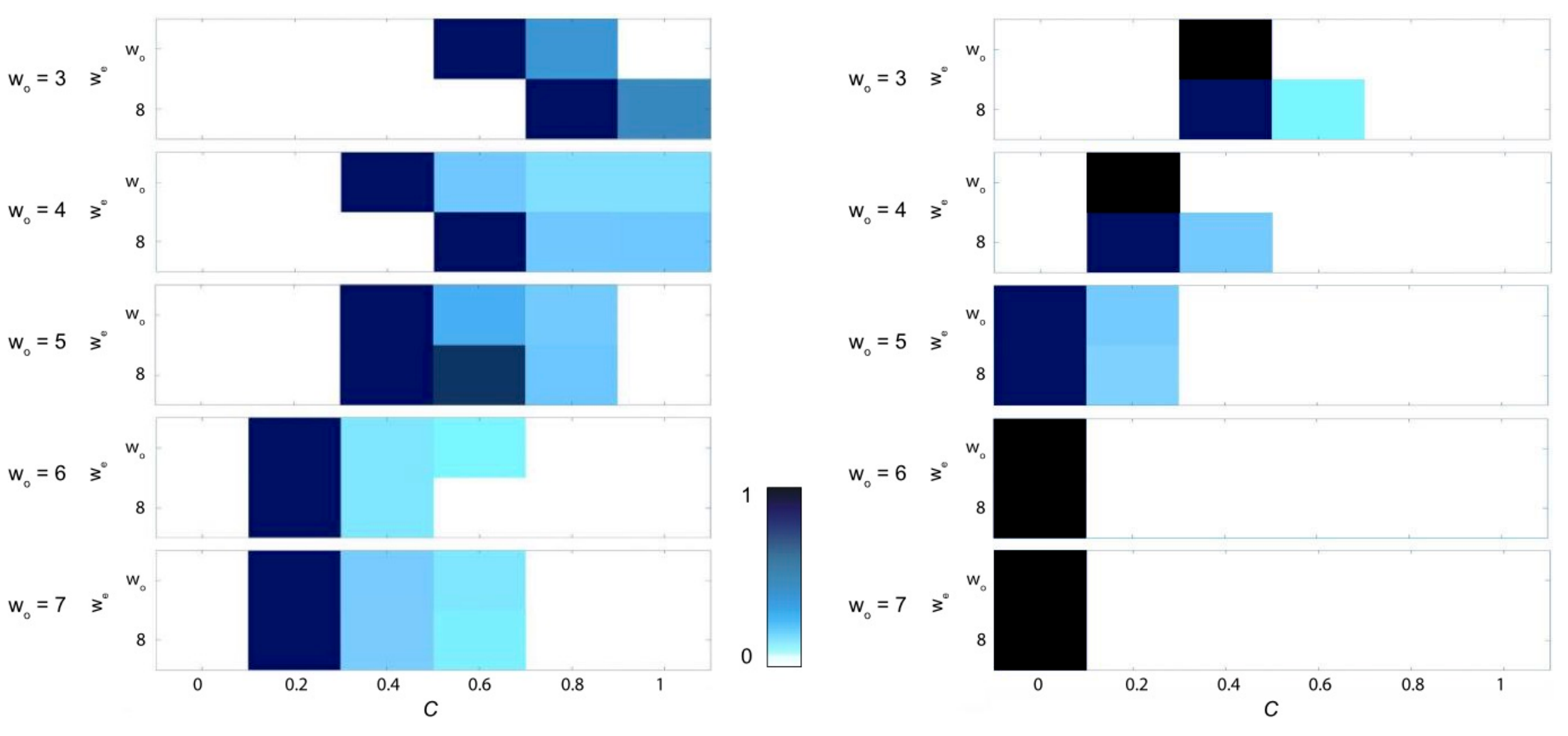}
    \caption{Heat maps of compression $C$ for all sampled trajectories of the states of $o$ (left), and for OEE trajectories only (right) shown for Case I and Case II CA. From top to bottom $w_o  = 3, 4, 5, 6$ and $7$.  For each $w_o$ shown are the distributions of $C$ for Case I CA for $w_e = w_o$ (top row in each panel) and for Case II CA with $w_e = 8$ (bottom row in each panel). Distributions are normalized to the total size of sampled trajectories for each $w_o$ and $w_e$  for each CA variant (see statistics in Tables \ref{compstats1} and \ref{compstats2}).}
    \label{fig:C}
\end{figure*}

\begin{figure*}[h]
    \centering
    \includegraphics[width=\textwidth]{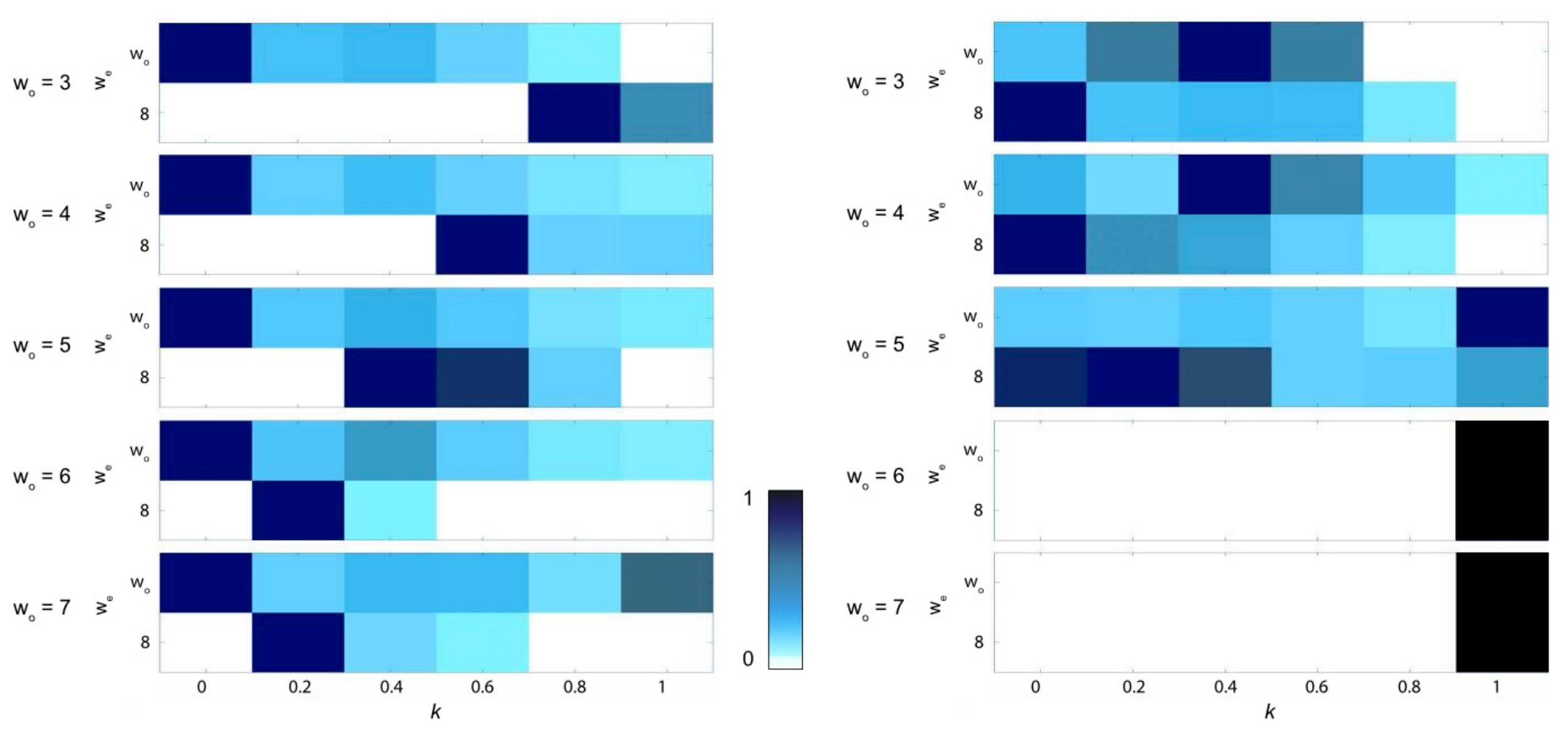}
    \caption{Heat maps of Lyapunov exponent $k$ for all sampled trajectories of the states of $o$ (left), and for OEE trajectories only (right) shown for Case I and Case II CA. From top to bottom $w_o  = 3, 4, 5, 6$ and $7$.  For each $w_o$ shown are the distributions of $k$ for Case I CA for $w_e = w_o$ (top row in each panel) and for Case II CA with $w_e = 8$ (bottom row in each panel). Distributions are normalized to the total size of sampled trajectories for each $w_o$ and $w_e$  for each CA variant (see statistics in Tables \ref{compstats1} and \ref{compstats2}).}
    \label{fig:k}
\end{figure*}

Calculated values for compressibility ($C$) and Lyapunov exponent ($k$), as defined in Section~\ref{sec:calc}, are compared for Case I and Case II CA in Figures \ref{fig:C} and \ref{fig:k}. Case III is not considered as the long-term dynamics display low complexity for the oscillatory attractor of the homogenous all-'0' and all-'1' states. For both Case I and Case II CA variants, increasing $w_o$ yields more OEE cases with lower $C$ values, such that larger ``organisms'' are more complex (Figure \ref{fig:C}). 

Case II CA yield lower $C$ values than Case I for the data shown as a result of the difference in the normalization implemented in Eq. \ref{eqn: C}, which for Case II CA is lower since the the width of $u$ is $w = w_o + 8$ for all $w_o$ explored, whereas for Case I CA the width of $u$ is $w = 2*w_o$ (for $w_e = w_o$ as shown). Additionally, as noted Case I is scalable as $w_e$ can be increased to generate higher complexity (lower $C$) cases. The Lyapunov exponent for Case II is in general higher than for Case I, indicating greater sensitivity to perturbations in the initial condition for Case II CA than Case I. For both CA variants, $k$ increases with increasing organism size $k$, such that larger organisms are more complex. 

\newpage

\newpage
\subsection*{Larger Systems} \label{sec:largeO}

\begin{figure*}[h]
    \centering
    \includegraphics[width=1\textwidth]{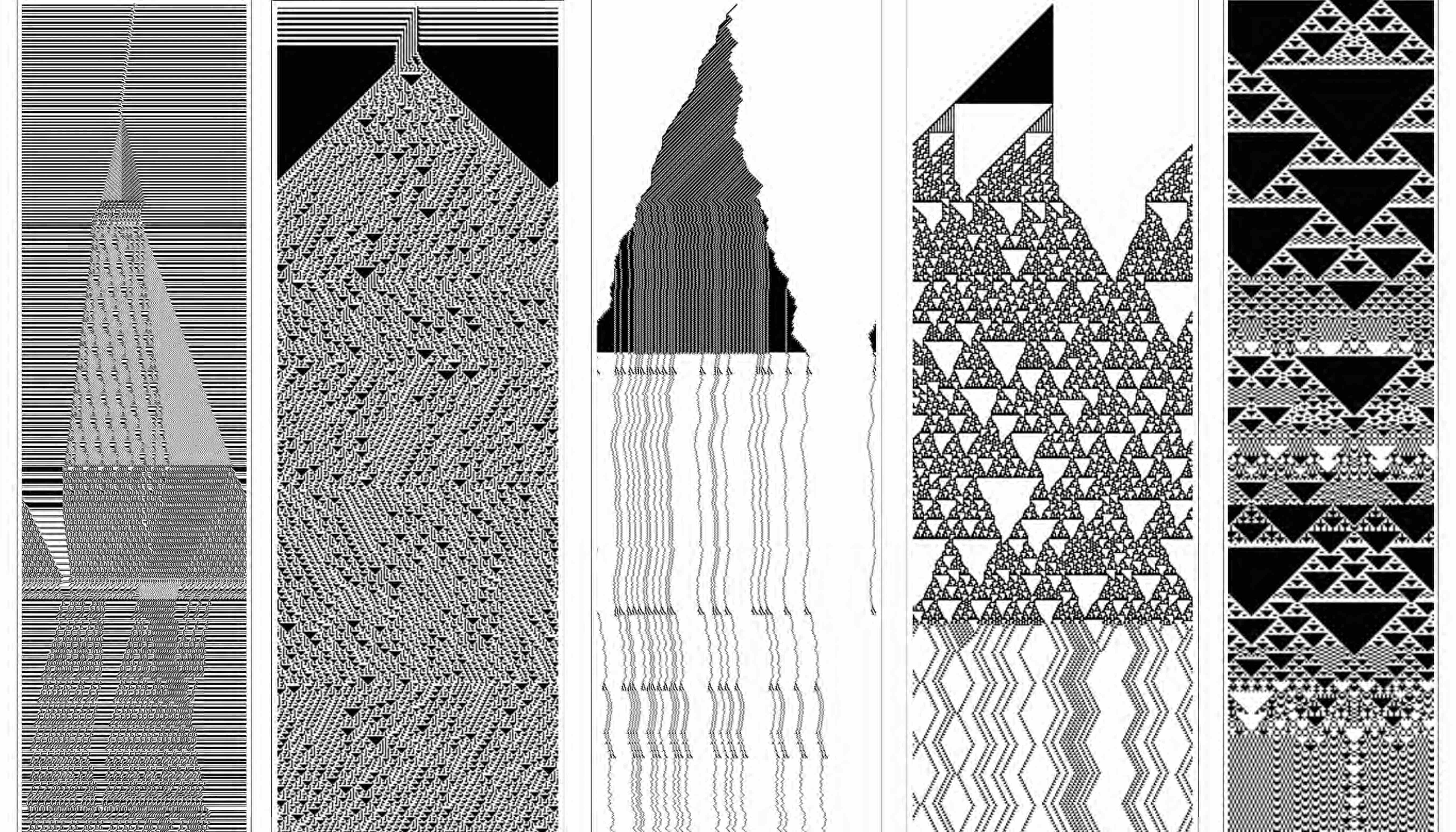}
    \caption{Example executions of the state trajectory of $o$ for Case I CA for large system size $w_o = 101$, with $w_e = w_o$.}
    \label{biginno}
\end{figure*}

Figure \ref{biginno} shows example executions of Case I state-dependent CA for large organisms of width $w_o = 101$, which visually demonstrate that the novelty of the dynamics reported herein scale to large system sizes.

\bibliography{SciRep_refs}

\section*{Acknowledgements}

This project was made possible through support of a grant from the Templeton World Charity Foundation. The opinions expressed in this publication are those of the author(s) and do not necessarily reflect the views of Templeton World Charity Foundation. We thank S. Wolfram, T. Rowland, T. Beynon, A. Rueda-Toicen, and C. Marletto, and the Emergence group at ASU for helpful discussions related to this work.

\section*{Author contributions}

A.A., H.Z. and S.I.W. designed the research; A.A. and S.I.W. performed research; A.A., H.Z., P.D. and S.I.W. analyzed data; A.A., H.Z., P.D. and S.I.W. wrote the paper. All data and related code has been made publically available at \url{https://github.com/alyssa-adams/OEE_Project}.

\section*{Competing financial interests}
The author(s) declare no competing financial interests.

\end{document}